\documentclass[11pt]{article}

\usepackage[preprint]{acl}

\usepackage{times}
\usepackage{latexsym}

\usepackage[T1]{fontenc}
\usepackage{textcomp}
\usepackage[utf8]{inputenc}

\usepackage{microtype}

\usepackage{inconsolata}

\usepackage{graphicx}
\usepackage{subcaption}  % For subfigures
\usepackage{booktabs}    % For professional tables
\usepackage{tabularx}    % For tabularx in include files
\usepackage{float}       % For [H] table placement
\usepackage{multirow}    % For multirow in tables
\usepackage{colortbl}    % For cell coloring
\usepackage{xcolor}      % For custom colors
\usepackage{soul}         % For highlighting with line breaks
\usepackage{tcolorbox}    % Highlighted example boxes
\usepackage{enumitem}
\usepackage{tikz}
\usepackage{pgfplots}
\pgfplotsset{compat=1.18}

\usepackage{pdfpages}
\usepackage{hyperref}

\definecolor{uzhblue}{RGB}{0,40,165}

\newcommand{\resourceicon}[1]{%
    \raisebox{-0.15em}{\includegraphics[height=0.95em]{figures/icons/#1}}%
}
\newcommand{\resourcelink}[2]{%
    \begingroup
    \def\UrlFont{\footnotesize\ttfamily}%
    \href{#1}{\nolinkurl{#2}}%
    \endgroup
}

\newcommand{\ExternalLink}{%
    \tikz[x=1.2ex, y=1.2ex, baseline=-0.05ex]{%
        \begin{scope}[x=1ex, y=1ex]
            \clip (-0.1,-0.1)
                --++ (-0, 1.2)
                --++ (0.6, 0)
                --++ (0, -0.6)
                --++ (0.6, 0)
                --++ (0, -1);
            \path[draw,
                line width = 0.5,
                rounded corners=0.5]
                (0,0) rectangle (1,1);
        \end{scope}
        \path[draw, line width = 0.5] (0.5, 0.5)
            -- (1, 1);
        \path[draw, line width = 0.5] (0.6, 1)
            -- (1, 1) -- (1, 0.6);
        }
    }

\title{Translation Asymmetry in LLMs as a Data Augmentation Factor: \\A Case Study for 6 Romansh Language Varieties}

\author{
  \textbf{Jannis Vamvas\textsuperscript{1}}\quad
  \textbf{Ignacio Pérez Prat\textsuperscript{2}}\quad
  \textbf{Angela Heldstab\textsuperscript{1}}\quad \\
  \textbf{Dominic P. Fischer\textsuperscript{1}}\quad
  \textbf{Sina Ahmadi\textsuperscript{1}}\quad
  \textbf{Rico Sennrich\textsuperscript{1}}
\\[0.3em]
  \textsuperscript{1}University of Zurich \quad
  \textsuperscript{2}Lia Rumantscha
\\
  \small{
    Correspondence: \href{mailto:vamvas@cl.uzh.ch}{vamvas@cl.uzh.ch}, \href{mailto:ignacio.perez.prat@rumantsch.ch}{ignacio.perez.prat@rumantsch.ch}
  }
}

\begin{document}
\maketitle
\begin{abstract}
Recent strategies for low-resource machine translation rely on LLMs to generate synthetic data based on text in higher-resource languages.
We revisit this idea for Romansh, a language with 6 distinct varieties.
LLMs tend to confuse these varieties when translating \textit{into} Romansh, but they are quite good at translating \textit{out of} Romansh into a high-resource language such as German.
Due to this asymmetry, the direction of data augmentation is a crucial choice.
We find that contrary to recent strategies, creating synthetic translations into the \textit{higher}-resource language is the superior approach, and only this approach allows us to surpass a Gemini~3 Pro baseline on German$\rightarrow$Romansh translation~(+23~BLEU over Gemini in the lowest-resource variety).
A human evaluation confirms that our experiments yield the first model that generates fluent translations in the individual Romansh varieties.
\end{abstract}

\vspace{0.5em}
\begin{tcolorbox}[
  colback=gray!12,
  colframe=gray!12,
  boxrule=0pt,
  arc=0pt,
  left=6pt, right=4pt, top=4pt, bottom=4pt,
  fontupper=\small,
]
{\renewcommand{\arraystretch}{1.35}
\begin{tabularx}{\linewidth}{@{}c@{\hspace{0.45em}}l@{\hspace{0.6em}}>{\raggedright\arraybackslash}X@{}}
\resourceicon{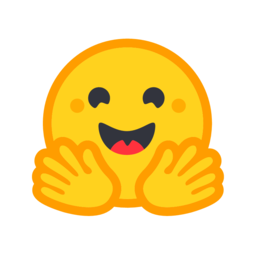} & Fine-tuned model & \resourcelink{https://hf.co/ZurichNLP/romansh-nllb-1.3b-ct2}{hf.co/ZurichNLP/romansh-nllb-1.3b-ct2} \\
\resourceicon{huggingface} & Back-translations & \resourcelink{https://hf.co/datasets/ZurichNLP/romansh-backtranslated}{hf.co/datasets/ZurichNLP/romansh-backtranslated} \\
\resourceicon{huggingface} & Human ratings & \resourcelink{https://hf.co/datasets/ZurichNLP/romansh-mt-evaluation}{hf.co/datasets/ZurichNLP/romansh-mt-evaluation} \\
\resourceicon{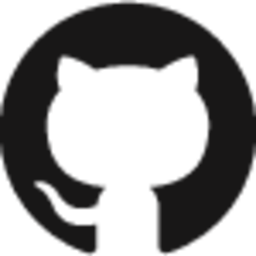} & Evaluation code & \resourcelink{https://github.com/ZurichNLP/romansh_mt_eval}{github.com/ZurichNLP/romansh_mt_eval} \\
\resourceicon{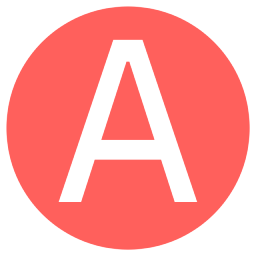} & Production system & \resourcelink{https://alas.liarumantscha.ch/}{alas.liarumantscha.ch} \\
\end{tabularx}}
\end{tcolorbox}
\vspace{0.4em}

\section{Introduction}
\begin{figure}
    \centering
    \includegraphics[width=\columnwidth]{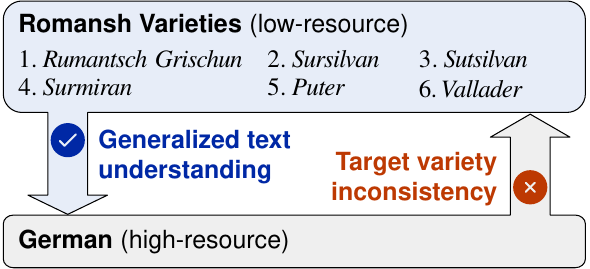}
    \caption{LLMs have asymmetric translation capabilities regarding low-resource or multi-variety languages like Romansh.
    In the case of Romansh, they demonstrate a general understanding of all varieties when translating \textit{out of} the language, but they fail to adhere to a specific target variety when translating \textit{into} the language.
    This asymmetry is relevant for data augmentation.
    }
    \label{fig:figure1}
\end{figure}

Traditionally, both forward translation and back-translation have been found to be helpful data augmentation strategies for neural MT (see,
 e.g., \citealp{burlot-yvon-2018-using}), and the choice of strategy depended on the available monolingual data.
In recent work, large-scale LLM-based forward translation from English was used to create synthetic training data for low-resource MT~\cite{de-gibert-etal-2025-scaling}.
 While this may be effective for some language pairs, we draw attention to the inherent asymmetry of LLM capabilities for low-resource or multi-variety languages~(Figure~\ref{fig:figure1}).

This paper studies \mbox{Romansh}, a minority language in Switzerland spoken by 40–60,000 speakers across several Alpine valleys.
Since Romansh has 6 distinct varieties, which are often mutually unintelligible, the development of variety-aware MT is a strong community need.
Romansh is also a prime example of translation asymmetry: LLMs are known to excel at understanding Romansh text in any variety when translating \textit{out of} Romansh, but struggle to produce text in the desired variety when translating \textit{into} Romansh~\cite{vamvas-et-al-2025-expanding, apertus2025apertusdemocratizingopencompliant}.
Therefore, generating synthetic Romansh text with LLMs might not be the optimal approach for data augmentation.

We present a controlled German$\leftrightarrow$Romansh case study that empirically compares the two directions of data augmentation: Either starting with text in the high-resource language and generating synthetic translations in the low-resource language (\textbf{\textit{HR$\rightarrow$LR augmentation}}), or investing into manual data collection in the low-resource language and then generating synthetic translations in the high-resource language~(\textbf{\textit{LR$\rightarrow$HR augmentation}}).

To compare these augmentation directions, we fine-tune NLLB~\cite{costa-jussa2024nllb} on the augmented data and evaluate German$\leftrightarrow$Romansh translation across the six varieties.
Surprisingly, we find that LR$\rightarrow$HR augmentation is the better approach for both downstream translation directions (German$\rightarrow$Romansh \textit{and} Romansh$\rightarrow$German).
Thus, for German$\rightarrow$Romansh MT, back-translating low-resource text outperforms forward-translating high-resource text. For Romansh$\rightarrow$German MT, forward-translating low-resource text is likewise superior to back-translating high-resource text.

\begin{figure}[t]
\centering
{\small
  \setlength{\tabcolsep}{8pt}
  \renewcommand{\arraystretch}{1.2}
  \begin{tabular}{@{}>{\columncolor[HTML]{E7E7E7}}>{\raggedright\arraybackslash}p{3.15cm}>{\columncolor[HTML]{EFEFEF}}>{\itshape}l@{}}
     & \\[-0.7em]
    Rumantsch~Grischun~(RG) & \textit{Quai n'hai jau betg ditg.} \\[0.1em]
    Sursilvan & \textit{Quei hai jeu buca detg.} \\[0.1em]
    Sutsilvan & \textit{Quegl ve jou betga getg.} \\[0.1em]
    Surmiran & \textit{Chegl vaia betg detg.} \\[0.1em]
    Puter & \textit{Que nu d'heja dit.} \\[0.1em]
    Vallader & \textit{Quai nun haja dit.} \\[0.4em]
    [German] & \textit{Das habe ich nicht gesagt.} \\[0.1em]
    [English] & \textit{I didn't say that.} \\[0.7em]
  \end{tabular}%
}
\caption{The varieties of Romansh are highly diverse, as shown in this example from the WMT24++ benchmark~\cite{deutsch-etal-2025-wmt24,vamvas-et-al-2025-expanding}.}
\label{fig:figure2}
\end{figure}

Our case study suggests that authentic data in low-resource languages remains highly beneficial for LLM-based data augmentation, and therefore that collecting monolingual text is still an important building block of low-resource MT. In summary, our main contributions are:
\begin{itemize}[topsep=0.4em, itemsep=0em]
    \item We provide a controlled comparison of LLM-based data augmentation in a low-resource scenario with 6 distinct target varieties.
    \item We validate prompting strategies such as few-shot prompting and dictionary prompting for data augmentation.
    \item We release our fine-tuned MT models, the synthetic back-translations used for LR$\rightarrow$HR augmentation, 10k+~human quality ratings, and the evaluation code.
\end{itemize}

\begin{figure}[t]
\centering
{\footnotesize
\captionsetup[subfigure]{skip=6pt, belowskip=0pt}
\begin{subfigure}{\columnwidth}
\centering

\begingroup
\setlength{\tabcolsep}{3pt}
\renewcommand{\arraystretch}{1.2}
\begin{tabularx}{0.84\linewidth}{@{}>{\raggedright\arraybackslash}p{0.9cm}|*{6}{>{\raggedleft\arraybackslash}X}@{}}
\multirow[b]{2}{*}{$\downarrow$~\textbf{tgt}}
& \multicolumn{6}{l}{$\textbf{ref}$~$\rightarrow$} \\
& RG & Surs.\ & Suts.\ & Surm.\ & Puter & Vall. \\
RG     & \cellcolor{uzhblue!40}39.8   & \cellcolor{uzhblue!14}18.1   & \cellcolor{uzhblue!3}8.7   & \cellcolor{uzhblue!7}11.9   & \cellcolor{uzhblue!7}11.9   & \cellcolor{uzhblue!11}15.4   \\
Surs.  & \cellcolor{uzhblue!22}25.2 & \cellcolor{uzhblue!28}29.5 & \cellcolor{uzhblue!2}8.4 & \cellcolor{uzhblue!5}10.4 & \cellcolor{uzhblue!4}9.7 & \cellcolor{uzhblue!7}12.2 \\
Suts.  & \cellcolor{uzhblue!25}27.2 & \cellcolor{uzhblue!13}17.5 & \cellcolor{uzhblue!4}10.0 & \cellcolor{uzhblue!7}12.2 & \cellcolor{uzhblue!6}11.0 & \cellcolor{uzhblue!9}13.5 \\
Surm.  & \cellcolor{uzhblue!16}19.6 & \cellcolor{uzhblue!8}13.0 & \cellcolor{uzhblue!4}9.3 & \cellcolor{uzhblue!14}17.8 & \cellcolor{uzhblue!5}10.7 & \cellcolor{uzhblue!6}11.7 \\
Puter  & \cellcolor{uzhblue!9}14.1   & \cellcolor{uzhblue!4}10.0   & \cellcolor{uzhblue!1}7.4   & \cellcolor{uzhblue!3}8.7   & \cellcolor{uzhblue!20}23.1   & \cellcolor{uzhblue!20}23.3   \\
Vall.  & \cellcolor{uzhblue!12}16.2 & \cellcolor{uzhblue!6}11.1 & \cellcolor{uzhblue!1}7.4 & \cellcolor{uzhblue!3}8.6 & \cellcolor{uzhblue!17}20.5 & \cellcolor{uzhblue!25}27.0 \\
\end{tabularx}
\endgroup

\caption{Gemini 2.5 Flash}
\label{subfig:main_confusion_gemini_25_flash}
\end{subfigure}

\vspace{1.2em}

\begin{subfigure}{\columnwidth}
\centering

\begingroup
\setlength{\tabcolsep}{3pt}
\renewcommand{\arraystretch}{1.2}
\begin{tabularx}{0.84\linewidth}{@{}>{\raggedright\arraybackslash}p{0.9cm}|*{6}{>{\raggedleft\arraybackslash}X}@{}}
\multirow[b]{2}{*}{$\downarrow$~\textbf{tgt}}
& \multicolumn{6}{l}{$\textbf{ref}$~$\rightarrow$} \\
& RG & Surs.\ & Suts.\ & Surm.\ & Puter & Vall. \\
RG     & \cellcolor{uzhblue!50}48.4   & \cellcolor{uzhblue!16}19.9   & \cellcolor{uzhblue!3}8.6   & \cellcolor{uzhblue!7}12.2   & \cellcolor{uzhblue!7}12.4   & \cellcolor{uzhblue!11}15.5   \\
Surs.  & \cellcolor{uzhblue!15}19.0 & \cellcolor{uzhblue!45}44.5 & \cellcolor{uzhblue!2}8.3 & \cellcolor{uzhblue!4}9.4 & \cellcolor{uzhblue!2}8.2 & \cellcolor{uzhblue!4}9.8 \\
Suts.  & \cellcolor{uzhblue!3}9.0 & \cellcolor{uzhblue!4}9.3 & \cellcolor{uzhblue!41}40.5 & \cellcolor{uzhblue!4}10.1 & \cellcolor{uzhblue!0}6.7 & \cellcolor{uzhblue!1}7.5 \\
Surm.  & \cellcolor{uzhblue!7}12.5 & \cellcolor{uzhblue!5}10.5 & \cellcolor{uzhblue!4}9.7 & \cellcolor{uzhblue!44}43.0 & \cellcolor{uzhblue!2}8.1 & \cellcolor{uzhblue!3}8.9 \\
Puter  & \cellcolor{uzhblue!7}11.9   & \cellcolor{uzhblue!3}8.9   & \cellcolor{uzhblue!0}6.3   & \cellcolor{uzhblue!2}8.0   & \cellcolor{uzhblue!46}44.9   & \cellcolor{uzhblue!22}25.0   \\
Vall.  & \cellcolor{uzhblue!10}14.8 & \cellcolor{uzhblue!4}10.1 & \cellcolor{uzhblue!0}6.7 & \cellcolor{uzhblue!2}8.1 & \cellcolor{uzhblue!22}24.7 & \cellcolor{uzhblue!45}44.6 \\
\end{tabularx}
\endgroup

\caption{Our system}
\label{subfig:main_confusion_nmt}
\end{subfigure}
}
\vspace{-0.45em}
\caption{When translating from German into Romansh, Gemini 2.5 Flash often produces the wrong target variety, which can be visualized through a confusion matrix that evaluates the system translations against the references for the individual varieties. Our fine-tuned NLLB model adheres to the target varieties and achieves higher BLEU.}
\label{fig:confusion-matrices-main}
\end{figure}

\section{Background: Language Varieties of Romansh}
Romansh, one of the four official languages of Switzerland, is a minority language spoken in only one Swiss canton (Grisons) and is considered endangered.
It has been argued that it is inaccurate to call Romansh a single language~\cite{caviezel1993geschichte} and that it is instead a continuum with no overarching language variety~\cite{liver_ratoromanisch_2010,grunert2024raetoromanisch}.
Within Grisons, the Romansh-speaking regions can be separated into five territories, where each of the five has their own written tradition, called `idiom'. An idiom is thus a standardized version of Romansh language varieties spoken within one of the Romansh-speaking regions, with its own orthography (see Figure~\ref{fig:figure2}) and its own literary tradition.
This split standardization dates back to the 1500s with religious and literary publications and was emphasized due to the lack of a centralized and common cultural space~\cite{grunert2024raetoromanisch}.

The largest speaker groups are Sursilvan (est.~50\% of Romansh speakers) and Vallader (20\%), while Sutsilvan has very few speakers (3\%). The other two are Surmiran (10\%) and Puter (12\%) \cite{furer2005aktuelle}. This distribution is broadly reflected in the available data (see Appendices~\ref{app:datasets} and \ref{app:training-data-statistics}).

% The efforts to introduce a common Romansh language date back to 1827 with none of them showing any success \cite{haiman_rhaeto-romance_1992, grunert2024raetoromanisch}, until 1982 with the introduction of Rumantsch Grischun.
Rumantsch Grischun~(RG), unlike the other five varieties, is not an idiom that was standardized based on spoken language, but rather a constructed language mostly based on Sursilvan, Surmiran, and Vallader \cite{liver_ratoromanisch_2010}. It was intended for governmental and institutional communication and explicitly introduced as complementary to the existing idioms \cite{coray_von_2008}. % to be applied where a supra-regional variety was needed .
The canton and the federal government use RG for official publications and communications.
%, and official media outlets, especially RTR, use it as well.
%There were many other approaches with the intention of promoting RG more heavily, which have however faced considerable resistance from the Romansh-speaking community~\cite{coray_von_2008}.
Most web data readily available is in RG, and for this reason, prior work on MT for Romansh has focused on RG only~(e.g., \citealp{muller-etal-2020-domain,NEURIPS2023_d49042a5}, as well as a commercial system by Supertext).
However, Romansh speakers use their native idioms both in school and at work, and therefore, most speakers have passive knowledge of RG at best.
Variety-aware MT would allow Romansh speakers a greater degree of access to content produced in other languages or other Romansh varieties.

\section{LLM Translation Asymmetry}
It is well known that LLMs perform better with low-resource languages as a source than as a target~(e.g., \citealp{bawden-yvon-2023-investigating,zhu-etal-2024-multilingual,omnilingualmtteam2026omnilingualmtmachinetranslation}).
Recent evaluations have shown the same asymmetry for Romansh~\cite{vamvas-et-al-2025-expanding,apertus2025apertusdemocratizingopencompliant}, and Table~\ref{tab:validation-backtranslation} reproduces this finding with four LLMs.

There are several potential explanations for the asymmetry.
One explanation is \textit{transfer asymmetry}: Knowledge transfer from related, higher-resource languages may compensate for limited low-resource language exposure more effectively for understanding, rather than generation.
An alternative explanation is \textit{standardization asymmetry}: Some low-resource languages might have more diverse variation than high-resource languages (relative to their representation in the training data), which could make the task of translating into these languages more ambiguous for a general-purpose LLM.
Finally, \textit{tokenization disparity} could mean that the generation of text in low-resource languages requires more decisions and that the errors and biases of autoregressive architectures~\cite{lesci-etal-2025-causal} compound more easily.

While it remains open to what degree these speculative explanations apply to the German$\leftrightarrow$Romansh language pair, we assume that \textit{standardization asymmetry} is especially important.
The confusion matrix in Figure~\ref{subfig:main_confusion_gemini_25_flash} illustrates how LLMs struggle to produce a specific variety of Romansh on demand: when translating text from German into specific varieties such as Sutsilvan, Surmiran or Puter, Gemini 2.5 Flash achieves a \textit{lower} BLEU score under the reference translations in the desired variety than under the reference translations in other, higher-resource varieties.
Some more recent versions of Gemini are less confused, but still strongly favor higher-resource varieties, especially RG~(Appendix~\ref{app:confusion-matrices-all-systems}).

In this paper, we show that LLM translation asymmetry has direct implications for data augmentation.

\section{Data Augmentation Methods}

\subsection{Forward and Back-translation}
\textit{Forward translation} generates synthetic targets from monolingual source text, while \textit{back-translation} generates synthetic sources from monolingual target text.
Comparative studies have found both methods effective for NMT~\cite{burlot-yvon-2018-using,bogoychev2020domaintranslationesenoisesynthetic}, and both remain widely used~\cite{kocmi-etal-2025-findings}.
Recently, \citet{de-gibert-etal-2025-scaling} found that large-scale forward translation with GPT-4o improves NMT for 7~low-resource languages, such as Basque and Georgian.
\citet{frontull-moser-2024-rule} combined forward and back-translation in an iterative process for Italian$\leftrightarrow$Ladin with GPT~3.5 back-translation as a seed.
However, there exists no systematic comparison of LLM-based forward and back-translation.

In this paper, we perform a comparison in a controlled setting for German$\leftrightarrow$Romansh MT, fixing the LLM (Gemini 2.5 Flash), prompt, and downstream NMT model~(NLLB).
We train our model variants jointly in both directions and on all varieties, using each sample bidirectionally.
While this multilingual setting differs from traditional bilingual back-translation~\cite{sennrich-etal-2016-improving}, it is more in line with our production requirements.

As a consequence, all synthetic translations can be considered both forward translations and back-translations, depending on the direction in which the downstream model will be used, but the open question is whether they should be created in the low-resource language (\textit{HR$\rightarrow$LR augmentation}, like \citealp{de-gibert-etal-2025-scaling}), or whether they should be created in the high-resource language based on text in the low-resource language~(\textit{LR$\rightarrow$HR augmentation}).
We will use these two terms in the next sections to avoid any ambiguity.

\subsection{Dictionary Prompting}\label{subsec:dictionary-prompting}
We experiment with a variant of LLM-based augmentation that appends bilingual dictionary entries to the prompt~\cite{ghazvininejad2023dictionarybasedphraselevelpromptinglarge,court-elsner-2024-shortcomings}.
Due to the high cost of this style of prompting, we test it only in the LR$\rightarrow$HR augmentation direction.

The dictionary information consists of the Romansh lemma for each word in the input document, together with potential German word-level translations and morphological analyses of the words, where available. For example, the Sursilvan word \textit{stos} is explained to the LLM as follows:

\makeatletter
\ifacl@linenumbers
  \nolinenumbers
\fi
\makeatother
\begin{tcolorbox}[colback=gray!20,colframe=gray!20,boxrule=0pt,arc=0pt,left=2pt,right=2pt,top=2pt,bottom=2pt,fontupper=\footnotesize]
\texttt{stos (lemma: stuer) \textrightarrow\ müssen}\\ \texttt{[PoS=V;Mood=IND;Tense=PRS;Person=2;Number=SG]}
\end{tcolorbox}
\makeatletter
\ifacl@linenumbers
  \linenumbers
\fi
\makeatother

\noindent{}Appendix~\ref{app:dictionary-prompt-example} presents a complete example prompt.

\section{Experimental Setup}

\subsection{Evaluation Data}
For evaluation, we primarily use the Romansh version of the WMT24++ benchmark~\cite{vamvas-et-al-2025-expanding}, paired with the German version from \citet{deutsch-etal-2025-wmt24}.
This benchmark contains 998 text segments from 4 domains (literary, news, social, speech).
We split the benchmark into a validation split (first 50\% documents per domain) and a test split (remaining 50\%), to allow for hyperparameter optimization and prompt engineering based on the validation split.
The final results are based on the test split.

We additionally evaluate on the more recent BOUQuET benchmark~\cite{andrews-etal-2025-bouquet}, which provides 854 German--Romansh sentence pairs covering 8 domains and combinations of 12 linguistic registers.
This benchmark is only available for the RG variety of Romansh.

\subsection{LLM-based Data Augmentation}
For creating the synthetic training data, we prompt the LLM with the instruction \textit{``Translate the following text into \{language\},''} translating segments of up to 500 tokens in a single request. The full prompts are reported in Appendix~\ref{sec:prompt-templates}.
We use greedy decoding for the data augmentation, leaving sampling-based augmentation~\cite{edunov-etal-2018-understanding} to future work.

Furthermore, results obtained on our validation split of WMT24++~(Table~\ref{tab:validation-backtranslation}) lead us to make the following choices:

\begin{itemize}
    \item \textbf{Gemini 2.5 Flash}~\cite{comanici2025gemini25pushingfrontier} achieves the best scores in both translation directions compared to three other LLMs of similar cost, confirming prior findings by~\citet{vamvas-et-al-2025-expanding}. Thus, we use this model for further experiments.
    \item \textbf{Disabling reasoning} allows us to save cost for the large-scale data augmentation experiments. Enabling reasoning (with a budget of 2048 tokens) only leads to modest gains on the validation set (+0.5 BLEU for German$\rightarrow$Romansh; +0.1 COMET for Romansh$\rightarrow$German).
    \item \textbf{Few-shot examples} are crucial for instruction following (+3.2 BLEU for German$\rightarrow$Romansh; +5.8 COMET for Romansh$\rightarrow$German). Zero-shot prompting without any examples results in translations into English instead of German, leading to a very poor BLEU score of 3.2.
    \item \textbf{Dictionary prompting} shows a positive effect (+0.4 COMET). Even though the effect is relatively small in terms of COMET, we still include this as a variant in the large-scale data augmentation experiment and the human evaluation, since the technique might affect long-tail phenomena that are harder to quantify with automatic metrics.
\end{itemize}

\begin{table}[t]
\centering
\footnotesize
\begin{tabularx}{\columnwidth}{@{}Xrr@{}}
\toprule
& \textbf{DE$\rightarrow$RM} & \textbf{RM$\rightarrow$DE} \\
\cmidrule(lr){2-2} \cmidrule(l){3-3}
\textbf{Approach} & \textbf{BLEU} & \textbf{BLEU / COMET} \\
\midrule
\mbox{\textit{Baseline LLMs}} & & \\[0.2em]
\mbox{Llama 3.3 (70B)} & 11.1 & 27.1 / 70.1 \\
\mbox{Apertus (70B)} & 13.7 & 35.6 / 84.3 \\
\mbox{GPT-4o} & 22.7 & 48.3 / 89.4 \\
\midrule
\mbox{\textit{Gemini 2.5 Flash}} & & \\[0.2em]
\mbox{Baseline prompt (\ref{app:forward-backtranslation})} & 23.6 & 51.2 / 91.0 \\
\mbox{\quad – with reasoning} & 24.1 & 50.1 / 91.1 \\
\mbox{\quad -- without few-shot examples} & 20.4 & 3.2 / 85.2 \\[0.2em]
\mbox{Dictionary prompting (\ref{app:backtranslation-dict})} & - & 51.4 / 91.4 \\
\mbox{\quad – with reasoning} & - & 49.4 / 91.5 \\
\bottomrule
\end{tabularx}

\caption{Validation results for multiple LLMs and data augmentation strategies, based on our validation split of the WMT24++ benchmark and averaged across the six varieties of Romansh. The results for baseline LLMs are reported for comparison, and only Gemini 2.5 Flash is used for the experiments below.
}
\label{tab:validation-backtranslation}
\end{table}

\noindent{}In total, we run four large-scale data augmentation experiments with Gemini 2.5 Flash (different data mixes, with or without dictionary prompting), with a total cost of approximately 2,400 USD~(Appendix~\ref{app:llm-requests-stats}).

\begin{table*}[t]
\centering

\footnotesize
\begin{tabularx}{\textwidth}{@{}Xr@{\hspace{2.5em}}rrr@{}}
\toprule
Variety & \# Raters & \# Fluency ratings & \# Document accuracy ratings & \# Segment accuracy preferences \\
\midrule
RG & 2 & 1,117~\textcolor{gray}{\phantom{6}(51 overlap)} & 168~\textcolor{gray}{(16 overlap)} & 531~\textcolor{gray}{\phantom{1,}(159 overlap)} \\
Sursilvan & 5 & 1,015~\textcolor{gray}{\phantom{6}(64 overlap)} & 204~\textcolor{gray}{(21 overlap)} & 684~\textcolor{gray}{\phantom{1,}(271 overlap)} \\
Sutsilvan & 3 & 987~\textcolor{gray}{(139 overlap)} & 166~\textcolor{gray}{(37 overlap)} & 704~\textcolor{gray}{\phantom{1,}(341 overlap)} \\
Surmiran & 2 & 1,017~\textcolor{gray}{(238 overlap)} & 168~\textcolor{gray}{(23 overlap)} & 668~\textcolor{gray}{\phantom{1,}(303 overlap)} \\
Puter & 3 & 929~\textcolor{gray}{(160 overlap)} & 158~\textcolor{gray}{\phantom{9}(0 overlap)} & 425~\textcolor{gray}{\phantom{1,11}(0 overlap)} \\
Vallader & 2 & 827~\textcolor{gray}{\phantom{65}(0 overlap)} & 166~\textcolor{gray}{\phantom{9}(0 overlap)} & 484~\textcolor{gray}{\phantom{1,11}(0 overlap)} \\
\midrule
\textbf{Total} & \textbf{17} & \textbf{5,892}~\phantom{(652 overlap)} & \textbf{1,030}~\phantom{(97 overlap)} & \textbf{3,496}~\phantom{(1,116 overlap)} \\
\bottomrule
\end{tabularx}
\normalsize

\caption{Human evaluation statistics: number of raters and annotations by target variety. The overlap counts in parentheses refer to items annotated by more than one rater, which we use for inter-rater agreement analysis.}
\label{tab:human-evaluation-aggregate-statistics}
\end{table*}

\begin{table*}[t]
\centering
\footnotesize
\begin{tabularx}{\textwidth}{@{}Xrrr@{}}
\toprule
\textbf{Metric} & Fluency & Document accuracy & Segment accuracy \\
\midrule
Item-level Pearson correlation between raters & 67.6 & 31.7 & 42.3 \\
Item-level pairwise agreement rate between raters & 68.5 & 59.0 & 57.8 \\
System-level Spearman correlation between raters & 92.0 & 75.0 & 90.0 \\
\bottomrule
\end{tabularx}
\normalsize

\caption{Inter-rater agreement metrics for the human evaluation, averaged across the 4–5 target varieties in which overlapping annotations have been collected.}
\label{tab:inter-rater-agreement-averaged}
\end{table*}

\subsection{Training Data}
As a prerequisite for the experiment, we compile monolingual and parallel data for the Romansh language varieties~(Appendix~\ref{app:datasets}).

For LR$\rightarrow$HR augmentation, we use 140M tokens of authentic monolingual Romansh data, mostly from web corpora, news and schoolbooks~(Table~\ref{tab:monolingual-romansh-data}).
In total, we include data from 12 different sources, many of which are already publicly available, or have been made available by us as part of this publication to enable future reproductions.

For HR$\rightarrow$LR augmentation from German, we compile two different datasets of an equivalent size:

\begin{itemize}
\item \textbf{Europarl:} In this variant, we follow \citet{de-gibert-etal-2025-scaling} and sample from German Europarl~\cite{koehn-2005-europarl}. This approach is directly comparable to prior work but is based on text in a different domain than the authentic Romansh data used for LR$\rightarrow$HR augmentation, which could confound the comparison.
\item \textbf{Newscrawl / FineWeb2:} This variant uses a German data mix that is comparable to the authentic Romansh data: 50\% news text via Newscrawl 2023~\cite{kocmi-etal-2025-findings} and 50\% web text via FineWeb2~\cite{penedo2025fineweb2pipelinescale}. Figure~\ref{fig:data-composition} shows that this data mix closely approximates the domain composition used for LR$\rightarrow$HR augmentation.
\end{itemize}

\noindent{}In all cases, we add 23M tokens of authentic parallel data, including around 3M word forms from Romansh–German bilingual dictionaries.
The LLM generates synthetic translations at the document level, which we then split into smaller segments for NMT fine-tuning.
We re-label the variety of all Romansh segments (including synthetic text) using an LID classifier for Romansh varieties~\cite{model-2025-rumantsch-idiom}.
Further preprocessing details and final training data statistics are in Appendices~\ref{app:preprocessing} and~\ref{app:training-data-statistics}, showing that comparable amounts of training data are used for all approaches.

\subsection{NMT Fine-tuning}
We fine-tune the NLLB-200-Distilled 1.3B model~\cite{costa-jussa2024nllb}, a multilingual NMT model with a transformer-based encoder-decoder architecture.
Since NLLB does not include Romansh as a pre-trained language, we extend the model vocabulary with~6 new language tokens corresponding to each Romansh variety.
We use temperature sampling with $T=1.5$ to upsample underrepresented translation directions.
Other hyperparameters are reported in Appendix~\ref{app:hyperparameters}.

\begin{table*}[t]
\centering

{\footnotesize
\begin{tabular*}{\textwidth}{@{\extracolsep{\fill}}lrrrrrrcr@{}}
\toprule
\mbox{\textbf{System}} & \multicolumn{6}{c}{\textbf{WMT24++}} & \textbf{BOUQuET} & \multirow{2}{*}{\textbf{Avg.}} \\
\cmidrule(lr){2-7} \cmidrule(lr){8-8} \cmidrule(lr){9-9}
 & \textbf{RG} & \textbf{Sursilvan} & \textbf{Sutsilvan} & \textbf{Surmiran} & \textbf{Puter} & \textbf{Vallader} & \textbf{RG} \\
\midrule
\mbox{Gemini 2.5 Flash} & 40.3 & 28.0 & 10.6 & 16.9 & 22.0 & 27.3 & 27.2 & 24.6 \\
\mbox{Gemini 3 Pro (preview)} & 45.3 & 37.1 & 17.3 & 27.1 & 34.7 & 36.1 & 31.3 & 32.7 \\
\mbox{Gemini 3.7 Flash} & 40.9 & 34.2 & 14.8 & 22.1 & 30.4 & 29.8 & 29.8 & 28.9 \\
\midrule
\mbox{\textit{Fine-tuned NLLB}} & & & & & & & & \\[0.35em]
\mbox{No data augmentation} & 35.9 & 29.4 & 29.2 & 25.3 & 31.0 & 26.0 & 27.9 & 29.2 \\[0.45em]
\mbox{HR$\rightarrow$LR augmentation:} & & & & & & & & \\[0.32em]
\mbox{– Europarl data} & 40.8 & 31.3 & 12.4 & 18.7 & 23.8 & 30.7 & 29.6 & 26.8 \\
\mbox{– Newscrawl / FineWeb2} & 41.4 & 30.4 & 12.1 & 18.5 & 24.6 & 29.4 & 29.7 & 26.6 \\[0.45em]
\mbox{LR$\rightarrow$HR augmentation:} & & & & & & & & \\[0.32em]
\mbox{– Baseline prompt} & 48.1 & 44.3 & \textbf{\underline{40.6}} & 42.3 & 44.8 & 44.3 & \textbf{\underline{33.1}} & 42.5 \\
\mbox{– Dictionary prompting} & \textbf{\underline{48.4}} & \textbf{\underline{44.5}} & 40.5 & \textbf{\underline{43.0}} & \textbf{\underline{44.9}} & \textbf{\underline{44.6}} & 32.6 & \textbf{\underline{42.6}} \\
\bottomrule
\end{tabular*}
}

\caption{\textbf{BLEU scores for German$\rightarrow$Romansh} translation on the second half of the WMT24++ benchmark, together with BOUQuET results for Rumantsch Grischun. Bold denotes the best system; \underline{underline} indicates the best variant among the fine-tuned NLLB models; the last column reports the arithmetic mean of the seven columns.}
\label{tab:detailed-de-to-rm-bleu}
\end{table*}

\begin{table*}[t]
\centering

{\footnotesize
\begin{tabular*}{\textwidth}{@{\extracolsep{\fill}}lrrrrrrcr@{}}
\toprule
\mbox{\textbf{System}} & \multicolumn{6}{c}{\textbf{WMT24++}} & \textbf{BOUQuET} & \multirow{2}{*}{\textbf{Avg.}} \\
\cmidrule(lr){2-7} \cmidrule(lr){8-8} \cmidrule(lr){9-9}
 & \textbf{RG} & \textbf{Sursilvan} & \textbf{Sutsilvan} & \textbf{Surmiran} & \textbf{Puter} & \textbf{Vallader} & \textbf{RG} \\
\midrule
\mbox{Gemini 2.5 Flash} & 93.7 & 93.1 & 89.8 & 91.7 & 92.3 & 92.7 & 97.6 & 93.0 \\
\mbox{Gemini 3 Pro (preview)} & 93.8 & 93.8 & 92.5 & 93.4 & 93.0 & 93.8 & 97.8 & 94.0 \\
\mbox{Gemini 3.7 Flash} & \textbf{94.3} & \textbf{94.0} & \textbf{93.3} & \textbf{93.9} & \textbf{93.4} & \textbf{94.1} & \textbf{97.9} & \textbf{94.4} \\
\midrule
\mbox{\textit{Fine-tuned NLLB}} & & & & & & & & \\[0.35em]
\mbox{No data augmentation} & 81.9 & 80.9 & 80.2 & 79.1 & 79.0 & 79.3 & 93.3 & 82.0 \\[0.45em]
\mbox{HR$\rightarrow$LR augmentation:} & & & & & & & & \\[0.32em]
\mbox{– Europarl data} & 89.1 & 89.6 & 88.6 & 88.6 & 88.9 & 88.4 & 96.2 & 89.9 \\
\mbox{– Newscrawl / FineWeb2} & 91.7 & 90.9 & 89.8 & 89.5 & 90.7 & 90.1 & 97.0 & 91.4 \\[0.45em]
\mbox{LR$\rightarrow$HR augmentation:} & & & & & & & & \\[0.32em]
\mbox{– Baseline prompt} & 92.4 & 92.3 & 90.8 & 91.1 & \underline{91.6} & 91.3 & \underline{97.3} & 92.4 \\
\mbox{– Dictionary prompting} & \underline{92.5} & \underline{92.4} & \underline{91.1} & \underline{91.4} & \underline{91.6} & \underline{91.5} & \underline{97.3} & \underline{92.5} \\
\bottomrule
\end{tabular*}
}

\caption{\textbf{COMET scores for Romansh$\rightarrow$German} translation on the second half of the WMT24++ benchmark, together with BOUQuET results for Rumantsch Grischun. Bold denotes the best system; \underline{underline} indicates the best variant among the fine-tuned NLLB models; the last column reports the arithmetic mean of the seven columns.}
\label{tab:detailed-rm-to-de-comet}
\end{table*}

\subsection{Automatic Evaluation}
For German$\rightarrow$Romansh, we compute BLEU scores~\cite{papineni-etal-2002-bleu} using SacreBLEU~\cite{post-2018-call}\footnote{Version: \texttt{\#:1|c:mixed|e:no|tok:13a|s:exp|v:2.5.1}}, due to the lack of a trained metric for this target language (complemented with a human evaluation, see Section~\ref{subsec:human-evaluation} below).
For Romansh$\rightarrow$German, we primarily report xCOMET~\cite{guerreiro-etal-2024-xcomet}, replicating the setup of \citet{vamvas-et-al-2025-expanding}.
We apply the model version \texttt{XCOMET-XL}\footnote{\url{https://hf.co/Unbabel/XCOMET-XL}} in \textit{reference-only} mode, providing only the German translation and reference without the Romansh source, as Romansh was not included in xCOMET's training data.
Following~\citet{kocmi-etal-2024-findings}, xCOMET scores are macro-averaged across domains to account for varying segment granularities.

\subsection{Human Evaluation}\label{subsec:human-evaluation}
We recruited 18 native speakers of all six Romansh varieties to blindly compare the three best systems in the German$\rightarrow$Romansh direction, plus the human references.
Fluency was rated on the level of individual segments, while accuracy was rated on the document level; both on a 7-point scale.

For most varieties, a part of the items were rated by more than one rater~(Table~\ref{tab:human-evaluation-aggregate-statistics}).
This allows us to calculate inter-rater agreement metrics (Table~\ref{tab:inter-rater-agreement-averaged} reports the numbers on average across target varieties, and Appendix~\ref{app:inter-rater-agreement} provides a per-variety analysis.)
System-level Spearman correlations mostly range from 80 to 100 for both fluency and accuracy, while item-level Pearson correlations are moderate (52--82 for fluency, 17--50 for accuracy).
\footnote{For Vallader fluency and Sursilvan accuracy, we initially observed low IAA and, after consulting with the respective raters, excluded three annotation runs from the final dataset
%, as the raters may have interpreted the instructions differently than intended,
and collected new ratings for Sursilvan accuracy from a third and fourth individual.
Therefore, the final number of raters included in the results is 17.}

To calculate system scores from the human ratings, we z-normalize the scores per rater and average them across segments.
We use bootstrap resampling~\cite{koehn-2004-statistical} to compute 95\% confidence intervals.

\section{Results}

Tables~\ref{tab:detailed-de-to-rm-bleu} and~\ref{tab:detailed-rm-to-de-comet} report the results of the automatic evaluation for translation directions.
Figure~\ref{fig:results-human} visualizes the results of the human evaluation of the top three systems; detailed per-variety results are reported in Appendix~\ref{app:detailed-human-results}.

\paragraph{HR$\rightarrow$LR augmentation does not benefit the lowest-resource varieties.}
The benefit of HR$\rightarrow$LR augmentation can be measured by comparing a model that was only trained on natively parallel data (``No data augmentation'') with a model that was trained on both the native and the synthetic parallel data.
Even though the HR$\rightarrow$LR augmented model is trained on 14$\times$ more data~(Appendix~\ref{app:training-data-statistics}), it does not consistently outperform the baseline.
In the German$\rightarrow$Romansh direction~(Table~\ref{tab:detailed-de-to-rm-bleu}), the model does improve on the higher-resource varieties (Rumantsch Grischun, Sursilvan and Vallader), but is worse in the lower-resource varieties (Sutsilvan, Surmiran and Puter), with a decline of -17 BLEU in the case of Sutsilvan.
This suggests that the synthetic text generated through HR$\rightarrow$LR augmentation is not helpful for fine-tuning an MT system on these lower-resource varieties.
An analysis of the synthetic text in Section~\ref{sec:analysis-of-augmented-data} will corroborate this hypothesis.

\paragraph{HR diversity does not help HR$\rightarrow$LR.}
As expected, HR$\rightarrow$LR augmentation is beneficial in the Romansh$\rightarrow$German direction~(Table~\ref{tab:detailed-rm-to-de-comet}), with an average improvement over the no-augmentation baseline of 7.9–9.4 COMET.
Comparing the effect of the two source datasets (Europarl and Newscrawl/FineWeb2), we find that the latter dataset gives a consistent improvement in the Romansh$\rightarrow$German direction (+1.5 COMET on average), but has no consistent effect in the German$\rightarrow$Romansh direction (-0.2 BLEU).
Thus, the domain diversity of the source text appears to be a minor factor for HR$\rightarrow$LR augmentation.

\paragraph{LR$\rightarrow$HR augmentation is superior for both downstream MT directions.}
An MT model fine-tuned on LR$\rightarrow$HR synthetic data outperforms the models trained on HR$\rightarrow$LR synthetic data and those without any data augmentation.
This is true for both test-time translation directions and for all varieties of Romansh, with average gains of 15.8 BLEU for German$\rightarrow$Romansh and 1.1 COMET for Romansh$\rightarrow$German.

A comparison of COMET scores and BLEU scores for Romansh$\rightarrow$German~(Appendix~\ref{app:detailed-rm-to-de-bleu}) indicates that the bidirectionality of the improvement could be explained by an increased semantic accuracy of the parallel data: On average, HR$\rightarrow$LR-augmented models have a similar German BLEU score as LR$\rightarrow$HR-augmented models, but they are clearly worse in terms of COMET.
Since COMET is more likely to pick up long-tail semantic errors than BLEU, this phenomenon could explain the disagreement between the metrics.

Overall, these results suggest that LLM-based data augmentation should ideally be performed in the LR$\rightarrow$HR direction, based on authentic monolingual data in the low-resource language, and that work to collect such data will continue to contribute to MT quality.

\paragraph{LLM-based data augmentation enables variety-aware MT into Romansh.}

\begin{figure}[t]
\centering
% Axis width = column minus label column (text width + gap). TikZ font=\footnotesize; do not wrap in \resizebox.
\definecolor{forestplotbackground}{RGB}{245,245,245}% HTML whitesmoke #F5F5F5
\begin{tikzpicture}[font=\footnotesize]
\begin{axis}[
    width=\dimexpr\columnwidth-29mm\relax,
    height=4.8cm,
    scale only axis,
    xmin=-1.0, xmax=0.75,
    ymin=0.5, ymax=4.25,
    xtick={-1.0,-0.5,0,0.5},
    xlabel={$z$-normalized human quality rating},
    xlabel style={yshift=2pt},
    % Fourth row centre at $y=3.75$ (between tight $3.5$ and uniform $4$).
    ytick={1,2,3,3.75},
    yticklabels={
      {\mbox{Human reference}},
      {Gemini 3 Pro \mbox{(preview)}},
      {LR$\rightarrow$HR augmented NLLB},
      {\begin{minipage}[t]{2.4cm}\raggedright\leftskip=1em\relax\noindent + Dictionary prompting\end{minipage}},
    },
    yticklabel style={
      align=left,
      anchor=east,
      text width=2.45cm,
    },
    y dir=reverse,
    axis y line=left,
    axis x line=bottom,
    tick align=outside,
    tick pos=left,
    enlarge y limits=false,
    axis background/.append style={fill=forestplotbackground},
    xmajorgrids,
    grid style={dashed, gray!30},
    legend style={
        legend columns=1,
        font=\footnotesize,
        draw=black!55,
        line width=0.4pt,
        fill=white,
        fill opacity=1,
        draw opacity=1,
        inner xsep=4pt,
        inner ysep=2pt,
        outer sep=0pt,
        column sep=2pt,
        row sep=1pt,
        cells={anchor=west},
        legend cell align=left,
        every legend image column/.append style={column sep=2pt},
        at={(rel axis cs:0.06,1.03)},
        anchor=north west,
    },
    legend image post style={scale=0.9},
    clip=false,
]

% Midlines between Human--Gemini and Gemini--NLLB; extend left into label band (shift matches axis width reserve).
\draw[gray!65, line width=0.55pt] ([xshift=-29mm]axis cs:-1.0,1.5) -- (axis cs:0.75,1.5);
\draw[gray!65, line width=0.55pt] ([xshift=-29mm]axis cs:-1.0,2.5) -- (axis cs:0.75,2.5);

\addplot+[
    only marks,
    mark=*, mark size=2.2pt,
    color=orange!85!black,
    error bars/.cd, x dir=both, x explicit,
    error bar style={line width=0.7pt, gray!55},
    error mark options={rotate=90, mark size=3pt, line width=0.7pt, draw=gray!55, fill=gray!55},
] table[x=x, y=y, x error minus=em, x error plus=ep, col sep=comma] {
    x,     em,   ep,   y
    0.47,  0.04, 0.03, 0.86
   -0.82,  0.05, 0.05, 1.86
    0.14,  0.05, 0.04, 2.86
    0.17,  0.05, 0.05, 3.61
};
\addlegendentry{Fluency}

\addplot+[
    only marks,
    mark=square*, mark size=2pt,
    color=blue!70!black,
    error bars/.cd, x dir=both, x explicit,
    error bar style={line width=0.7pt, gray!55},
    error mark options={rotate=90, mark size=3pt, line width=0.7pt, draw=gray!55, fill=gray!55},
] table[x=x, y=y, x error minus=em, x error plus=ep, col sep=comma] {
    x,     em,   ep,   y
    0.54,  0.12, 0.12, 1.14
   -0.75,  0.12, 0.12, 2.14
   -0.00,  0.12, 0.12, 3.14
    0.16,  0.13, 0.14, 3.89
};
\addlegendentry{Accuracy}

\end{axis}
\end{tikzpicture}
\caption{Human ratings for German$\rightarrow$Romansh translation, averaged across the six target varieties. The error bars show 95\% bootstrap confidence intervals.}
\label{fig:results-human}
\end{figure}
\begin{figure*}[t]
\centering
\footnotesize
\definecolor{pieeuroparl}{RGB}{215, 215, 215}
\definecolor{piefineweb}{RGB}{48,98,255}
\definecolor{pienews}{RGB}{117, 150, 255}
\definecolor{pieschool}{RGB}{0,40,165}
\definecolor{pieother}{RGB}{0,20,82}
\definecolor{pievarrg}{RGB}{164,210,51}
\definecolor{pievarsurs}{RGB}{200,228,133}
\definecolor{pievarsuts}{RGB}{255,222,143}
\definecolor{pievarsurm}{RGB}{254,147,103}
\definecolor{pievarputer}{RGB}{243,83,127}
\definecolor{pievarvall}{RGB}{191,13,62}

\newcommand{\pielegenditem}[3]{%
  \begin{scope}[shift={#2}]
    \fill[#1, draw=black!35, line width=0.15pt] (0,0) rectangle ++(0.12,0.12);
    \node[anchor=west, font=\footnotesize] at (0.16,0.06) {#3};
  \end{scope}
}
\newcommand{\pielegendtitle}[2]{%
  \begin{scope}[shift={#1}]
    \node[anchor=west, font=\bfseries\footnotesize] at (0,0.06) {#2};
  \end{scope}
}
\newcommand{\pietitle}[1]{%
  \node[align=center, font=\footnotesize, text width=3.5cm] at (0,1.65) {#1};
}

\begin{tabular*}{\textwidth}{@{\extracolsep{\fill}}cccc@{}}
\begin{tikzpicture}[font=\scriptsize, line join=round]
  \def\r{1.05}
  \pietitle{\textbf{LR$\rightarrow$HR augmentation}\\authentic Romansh text}
  \path[fill=pienews, draw=white, line width=0.6pt]
    (0,0) -- (90:\r) arc[start angle=90, end angle=252.8, radius=\r] -- cycle;
  \path[fill=pieschool, draw=white, line width=0.6pt]
    (0,0) -- (252.8:\r) arc[start angle=252.8, end angle=264.7, radius=\r] -- cycle;
  \path[fill=pieother, draw=white, line width=0.6pt]
    (0,0) -- (264.7:\r) arc[start angle=264.7, end angle=269.5, radius=\r] -- cycle;
  \path[fill=piefineweb, draw=white, line width=0.6pt]
    (0,0) -- (269.5:\r) arc[start angle=269.5, end angle=450, radius=\r] -- cycle;
\end{tikzpicture}
&
\begin{tikzpicture}[font=\scriptsize, line join=round]
  \def\r{1.05}
  \pietitle{\textbf{HR$\rightarrow$LR augmentation}\\based on Europarl}
  \path[fill=pieeuroparl] (0,0) circle[radius=\r];
\end{tikzpicture}
&
\begin{tikzpicture}[font=\scriptsize, line join=round]
  \def\r{1.05}
  \pietitle{\textbf{HR$\rightarrow$LR augmentation}\\\mbox{based on Newscrawl / FineWeb2}}
  \path[fill=pienews, draw=white, line width=0.6pt]
    (0,0) -- (90:\r) arc[start angle=90, end angle=271.2, radius=\r] -- cycle;
  \path[fill=piefineweb, draw=white, line width=0.6pt]
    (0,0) -- (271.2:\r) arc[start angle=271.2, end angle=450, radius=\r] -- cycle;
\end{tikzpicture}
&
\begin{tikzpicture}[font=\footnotesize]
  \path[use as bounding box] (-0.05,-0.62) rectangle (2.85,1.35);
  \begin{scope}[shift={(-0.1,0.52)}]
    \pielegendtitle{(-0.15,0.74)}{Domains:}
    \pielegenditem{pieeuroparl}{(0,0.34)}{Europarl}
    \pielegenditem{pienews}{(0,0)}{News}
    \pielegenditem{piefineweb}{(0,-0.34)}{FineWeb2 and similar}
    \pielegenditem{pieschool}{(0,-0.68)}{Schoolbooks}
    \pielegenditem{pieother}{(0,-1.02)}{Other}
  \end{scope}
\end{tikzpicture}
\\[1.0em]
\begin{tikzpicture}[font=\scriptsize, line join=round]
  \def\r{1.05}
  \path[fill=pievarrg, draw=white, line width=0.6pt]
    (0,0) -- (90:\r) arc[start angle=90, end angle=279.5, radius=\r] -- cycle;
  \path[fill=pievarsurs, draw=white, line width=0.6pt]
    (0,0) -- (279.5:\r) arc[start angle=279.5, end angle=375.7, radius=\r] -- cycle;
  \path[fill=pievarsuts, draw=white, line width=0.6pt]
    (0,0) -- (375.7:\r) arc[start angle=375.7, end angle=384.0, radius=\r] -- cycle;
  \path[fill=pievarsurm, draw=white, line width=0.6pt]
    (0,0) -- (384.0:\r) arc[start angle=384.0, end angle=401.8, radius=\r] -- cycle;
  \path[fill=pievarputer, draw=white, line width=0.6pt]
    (0,0) -- (401.8:\r) arc[start angle=401.8, end angle=415.2, radius=\r] -- cycle;
  \path[fill=pievarvall, draw=white, line width=0.6pt]
    (0,0) -- (415.2:\r) arc[start angle=415.2, end angle=450, radius=\r] -- cycle;
\end{tikzpicture}
&
\begin{tikzpicture}[font=\scriptsize, line join=round]
  \def\r{1.05}
  \path[fill=pievarrg, draw=white, line width=0.6pt]
    (0,0) -- (90:\r) arc[start angle=90, end angle=219.9, radius=\r] -- cycle;
  \path[fill=pievarsurs, draw=white, line width=0.6pt]
    (0,0) -- (219.9:\r) arc[start angle=219.9, end angle=246.9, radius=\r] -- cycle;
  \path[fill=pievarsuts, draw=white, line width=0.6pt]
    (0,0) -- (246.9:\r) arc[start angle=246.9, end angle=256.3, radius=\r] -- cycle;
  \path[fill=pievarsurm, draw=white, line width=0.6pt]
    (0,0) -- (256.3:\r) arc[start angle=256.3, end angle=329.9, radius=\r] -- cycle;
  \path[fill=pievarputer, draw=white, line width=0.6pt]
    (0,0) -- (329.9:\r) arc[start angle=329.9, end angle=351.4, radius=\r] -- cycle;
  \path[fill=pievarvall, draw=white, line width=0.6pt]
    (0,0) -- (351.4:\r) arc[start angle=351.4, end angle=450, radius=\r] -- cycle;
\end{tikzpicture}
&
\begin{tikzpicture}[font=\scriptsize, line join=round]
  \def\r{1.05}
  \path[fill=pievarrg, draw=white, line width=0.6pt]
    (0,0) -- (90:\r) arc[start angle=90, end angle=251.3, radius=\r] -- cycle;
  \path[fill=pievarsurs, draw=white, line width=0.6pt]
    (0,0) -- (251.3:\r) arc[start angle=251.3, end angle=279.6, radius=\r] -- cycle;
  \path[fill=pievarsuts, draw=white, line width=0.6pt]
    (0,0) -- (279.6:\r) arc[start angle=279.6, end angle=291.7, radius=\r] -- cycle;
  \path[fill=pievarsurm, draw=white, line width=0.6pt]
    (0,0) -- (291.7:\r) arc[start angle=291.7, end angle=338.1, radius=\r] -- cycle;
  \path[fill=pievarputer, draw=white, line width=0.6pt]
    (0,0) -- (338.1:\r) arc[start angle=338.1, end angle=369.0, radius=\r] -- cycle;
  \path[fill=pievarvall, draw=white, line width=0.6pt]
    (0,0) -- (369.0:\r) arc[start angle=369.0, end angle=450, radius=\r] -- cycle;
\end{tikzpicture}
&
\begin{tikzpicture}[font=\footnotesize]
  \path[use as bounding box] (-0.05,-0.80) rectangle (2.85,1.35);
  \begin{scope}[shift={(-0.1,0.46)}]
    \pielegendtitle{(-0.15,0.74)}{Romansh varieties:}
    \pielegenditem{pievarrg}{(0,0.34)}{Rumantsch Grischun}
    \pielegenditem{pievarsurs}{(0,0)}{Sursilvan}
    \pielegenditem{pievarsuts}{(0,-0.34)}{Sutsilvan}
    \pielegenditem{pievarsurm}{(0,-0.68)}{Surmiran}
    \pielegenditem{pievarputer}{(0,-1.02)}{Puter}
    \pielegenditem{pievarvall}{(0,-1.36)}{Vallader}
  \end{scope}
\end{tikzpicture}
\end{tabular*}
\caption{Comparison of Romansh data (authentic / synthetic) with respect to the included domains~(top row) and the proportion of Romansh varieties according to LID classification~(bottom row).}
\label{fig:data-composition}
\end{figure*}

\begin{table*}[t]
\centering
\footnotesize
\setlength{\tabcolsep}{3pt}
\begin{tabularx}{\textwidth}{@{}X@{\hspace{2.4em}}rrrrrrr@{\hspace{1.4em}}rrrrrrr@{}}
\toprule
\textbf{Dataset} & \multicolumn{7}{c}{\textbf{LID agreement}} & \multicolumn{7}{c}{\textbf{Dictionary coverage}} \\
\cmidrule(lr){2-8} \cmidrule(l){9-15}
 & RG & Surs. & Suts. & Surm. & Puter & Vall. & Avg. &
 RG & Surs. & Suts. & Surm. & Puter & Vall. & Avg. \\
\midrule
\mbox{Authentic Romansh text} & - & - & - & - & - & - & - & 90.5 & 87.9 & 88.5 & 87.7 & 83.9 & 87.1 & 87.6 \\
\midrule
\multicolumn{15}{@{}l}{\textit{HR$\rightarrow$LR synthetic datasets:}} \\[0.2em]
– based on Europarl & 100.0 & 40.3 & 17.3 & 80.9 & 33.1 & 95.8 & 61.2 & 93.5 & 80.3 & 58.9 & 71.5 & 77.9 & 83.0 & 77.5 \\
\mbox{– Newscrawl / FineWeb2} & 99.2 & 41.5 & 17.0 & 61.6 & 47.0 & 86.9 & 58.9 & 89.2 & 74.9 & 57.5 & 66.0 & 72.9 & 76.6 & 72.9 \\
\bottomrule
\end{tabularx}

\caption{Computational analysis of Romansh data (authentic / synthetic) with regard to adherence to variety-specific features.
\textbf{LID agreement} quantifies the proportion of synthetic samples where the requested target variety matches the prediction of an LID system.
\textbf{Dictionary~coverage} is the proportion of words in the text that overlap with variety-specific Romansh dictionaries.}
\label{tab:synthetic-dataset-analysis}
\end{table*}

Our fine-tuned NLLB model outperforms Gemini 3 Pro by 9.9 BLEU on average for German$\rightarrow$Romansh translation, with gains of up to 23.2 BLEU for the lowest-resource variety, Sutsilvan~(Table~\ref{tab:detailed-de-to-rm-bleu}).
The human evaluation~(Figure~\ref{fig:results-human}) shows that speakers rate our model substantially higher than Gemini in fluency. At the system level, they also prefer our model for accuracy, although item-level agreement on accuracy is only moderate~(Table~\ref{tab:inter-rater-agreement-averaged}).
In the Romansh$\rightarrow$German direction, our model still underperforms Gemini in terms of COMET, which underlines the asymmetry of Gemini's capabilities.

Including dictionary information in the data augmentation prompt did slightly improve the downstream model both in the automatic evaluation (+0.1 BLEU / +0.1 COMET on average) and the human evaluation, but below statistical significance and without consistency across the individual varieties.
Future work could employ the available dictionary information beyond prompting, for instance for rejection fine-tuning.

The low performance of HR$\rightarrow$LR augmentation for German$\rightarrow$Romansh translation raises the question what specific deficiencies there are in the synthesized Romansh text and how it compares to the authentic text that is used for LR$\rightarrow$HR augmentation.
We perform a computational analysis using a language identification~(LID) system~\cite{model-2025-rumantsch-idiom} and a dictionary look-up software that matches word forms in Romansh text with variety-specific dictionaries~\cite{fischer-et-al-rumlem}.

\section{Analysis of Augmented Data}\label{sec:analysis-of-augmented-data}

\paragraph{LLMs struggle to synthesize identifiable text in low-resource varieties.} The left-hand side of Table~\ref{tab:synthetic-dataset-analysis} shows that only 17\% of the synthetic Sutsilvan samples are identified as Sutsilvan by the LID system, and less than 50\% of synthetic Puter samples are identified as Puter.
This is consistent with the confusion matrix of Gemini 2.5 Flash in Figure~\ref{fig:confusion-matrices-main}, where translations into Sutsilvan and Puter score higher against RG and Vallader references than the references for the actual target varieties.

However, a potential confound of such an LID analysis is that LID might be inherently easier for some varieties. For example, the relatively high LID agreement of synthetic Surmiran (61.6–80.9\%) could be explained by the especially high recall of the LID classifier on Surmiran~\cite{model-2025-rumantsch-idiom}.

When training the MT system, we make sure to label the samples using the LID system,  overriding the LLM-produced label.
As a consequence, the HR$\rightarrow$LR augmented data are imbalanced with regard to the varieties~(Figure~\ref{fig:data-composition}), albeit to a lesser degree than the available authentic Romansh data.

\paragraph{LLMs struggle to generate variety-specific word forms.}
The right-hand side of Table~\ref{tab:synthetic-dataset-analysis} indicates that the synthetic Romansh text contains fewer words that match with a dictionary than the authentic Romansh text.
For example, only 57.5–58.9\% of words in synthetic Sutsilvan text can be found in a Sutsilvan dictionary, whereas 88.5\% of words in authentic Sutsilvan text can be identified.

\medskip{}

Taken together, the analysis indicates that HR$\rightarrow$LR augmented data are less useful for MT fine-tuning because they reflect the LLM's bias towards higher-resource varieties and limited exposure to the lower-resource varieties.

\section{Conclusion}
Our case study on six Romansh language varieties shows that translation direction matters when an LLM is used for data augmentation.
Back-translation from the lower-resource language produces better training signals than forward translation from the higher-resource language.
This suggests that collecting authentic monolingual data may remain essential in similar low-resource settings.

\section*{Limitations}
We limit our controlled data augmentation experiments to Gemini 2.5 Flash and greedy decoding.
Gemini 2.5 Pro performed slightly better than Flash in preliminary experiments, but using it for data augmentation with all available data would have exceeded our budget. Experiments with larger models and alternative sampling strategies thus remain future work.
Additionally, we did not test dictionary prompting with forward translation from German, due to the high cost of LLM usage for brute-force dictionary prompting~(Appendix~\ref{app:llm-requests-stats}).

We trained five models, two of which were included in the human evaluation.
Alternative data augmentation strategies were evaluated intrinsically by translating samples from the WMT24++ benchmark and computing COMET~(Table~\ref{tab:validation-backtranslation}), without training and evaluating a downstream MT model.

Our evaluation focuses on German$\rightarrow$Romansh and Romansh$\rightarrow$German translation.
Future work could also explore translation \textit{between} Romansh varieties, as well as data augmentation for these directions.
 The human evaluation study was limited to the German$\rightarrow$Romansh direction, since this is the most challenging direction and the direction for which no trained MT metrics are currently available.
Finally, a limitation of the study is that it does not analyze document-level errors in a targeted way, such as inconsistency in terminology and named entities. However, the effect size in Figure 4 makes it unlikely that document-level aspects would affect the overall ranking between the systems we compared.

\section*{Ethical considerations}
We release our trained model under a CC-BY-NC 4.0 license, in line with the original license of NLLB.
For training, we use a mix of publicly available data (links in Appendix~\ref{app:datasets}), and a number of datasets that were made available to us for research use (listed in Appendix~\ref{app:datasets} without link).

Regarding privacy, the web-crawled portion of our training data (FineWeb2 and FinePDFs) has email addresses and IP addresses anonymized by default.
The other datasets we used, such as news articles, schoolbooks, and dictionaries, are at low risk of containing personally identifiable information.

For the human evaluation, raters were personally recruited based on their native Romansh proficiency.
They were compensated at a standard hourly rate.
We release the human evaluation data without any personally identifiable information.

\section*{Acknowledgments}
We thank RTR and Fundaziun Patrimoni Cultural RTR for their support.
We are grateful to Zachary Hopton, Diana Merkle, Anna Rutkiewicz and Sudehsna Sivakumar for help with data curation, Uniun dals Grischs for contributing dictionary data for Puter and Vallader, and Giuanna Caviezel, Not Soliva and their seminar participants for helpful feedback.
We also acknowledge the contribution of the native speakers who participated in the human evaluation study, namely Claudia Cadruvi, Martina Caprez, Marietta Cathomas, Eliane Cathomen, Laura Decurtins, Andri Florineth, Arina Lazzarini, Viviana Lazzarini, Lea Livers, Patrick Meister, Gierina Michael, Bettina Nicca, Zegna Pittet-Dosch, Barbara Riesch, Manuela Schnoz-Flury, and Annalea Stuppan, and those who would rather not be named.
For this publication, use was made of media data made available via Swissdox\@LiRI by the Linguistic Research Infrastructure of the University of Zurich (see \url{https://www.liri.uzh.ch/en/services/swissdox.html} for more information).


\bibliography{custom}

\appendix
\vfill

\pagebreak

\section{NLLB Fine-tuning Hyperparameters}\label{app:hyperparameters}

\begin{table}[H]
\centering
\small
\begin{tabularx}{\columnwidth}{@{}lX@{}}
\toprule
\textbf{Hyperparameter} & \textbf{Value} \\
\midrule
Base model & NLLB-200-Distilled 1.3B \\
Learning rate & $2 \times 10^{-4}$ \\
Warmup updates & 150 \\
Max epochs & 10 \\
Stopping criterion & Validation loss on a held-out split of the training data \\
Optimizer & Adam ($\beta_1=0.9$, $\beta_2=0.98$, $\epsilon=10^{-6}$) \\
LR scheduler & Inverse square root \\
Dropout & 0.1 \\
Label smoothing & 0.1 \\
Gradient clipping & 1.0 \\
Weight decay & 0.01 \\
Mixed precision & bfloat16 \\
Max tokens per batch & 4096 \\
Update frequency & 8 \\
Effective batch size & 32,768 tokens \\
Max sequence length & 512 tokens \\
Hardware & Single H100 GPU (80GB memory) \\
\bottomrule
\end{tabularx}
\caption{Fine-tuning hyperparameters.}
\label{tab:hyperparameters}
\end{table}

\section{NLLB Decoding}\label{app:decoding}

For NMT inference, we use the CTranslate2 toolkit\footnote{\url{https://github.com/OpenNMT/CTranslate2}} with beam search (beam size 4) and a maximum decoding length of 256 tokens.
Input texts are segmented before translation: For German$\rightarrow$Romansh, we split German into sentences using spaCy with the \texttt{de\_dep\_news\_trf} model; for Romansh$\rightarrow$German, we split Romansh into text segments using SaT~\cite{frohmann-etal-2024-segment}, with model \texttt{sat-12l-sm} and a maximum segment length of 500 characters.

\section{Data Preprocessing}\label{app:preprocessing}

\paragraph{Monolingual Data}
Monolingual data were preprocessed on the document level before performing the LLM-based data augmentation.
We remove documents/segments with excessive punctuation (>50\% punctuation tokens), collapse repeated punctuation (max 3 consecutive), filter short documents (<5 tokens), split long documents (>500 tokens) along newline boundaries, and deduplicate.

\paragraph{Parallel Data}
Parallel data (both synthetic and natively parallel) underwent similar filtering: we remove pairs with excessive punctuation, collapse repeated punctuation, filter non-Romansh text using a dictionary-based heuristic (at least 50\% of the words should be found in a Romansh dictionary), and deduplicate.
We then segment the documents into lines based on newline boundaries, align them across languages using Vecalign~\cite{thompson-koehn-2019-vecalign} with embeddings from OpenAI's \texttt{text-embedding-3-small} model, and filter out pairs with a length ratio greater than 1.5.

\paragraph{Romansh Variety Classification}
We classify all Romansh text (including synthetic text generated through forward translation, but excluding the parallel data derived from dictionaries) into one of 6 Romansh varieties using a Support Vector Machine classifier released by~\citet{model-2025-rumantsch-idiom}.

\section{Details on Dictionary Prompting}\label{app:dictionary-prompting}

We use Romansh--German dictionaries for all six Romansh varieties.
For Vallader and Puter, we use dictionaries provided by Uniun dals Grischs.\footnote{\url{https://www.udg.ch/dicziunari}}
For the other varieties, we use dictionaries from the Pledari Grond project.\footnote{\url{https://pledarigrond.ch}}

For each word in the source text, we include the following information in the prompt: the Romansh word form as it appears in the text, the inferred lemma, all corresponding German lemmas, and all possible morphological analyses of the word form based on paradigms provided by the dictionaries.
We used the \texttt{rumlem} software~\cite{fischer-et-al-rumlem} to derive this information from the Romansh text.
We do not use context-sensitive lemmatization, which we defer to future work.
Similarly, we do not inflect the German lemmas.

For consistency, the few-shot examples in the dictionary prompting setup also include dictionary entries.

To reduce prompt length, we exclude dictionary information for high-frequency words, specifically those with a rank below 500 per variety in our monolingual datasets.

We also experimented with approaches requiring less dictionary information, such as providing relevant dictionary entries only in a follow-up prompt based on an unconstrained translation created without dictionary information~(cf.~\citealp{bogoychev-chen-2023-terminology}).
However, preliminary experiments indicated that simply providing all entries in the initial prompt works best in terms of BLEU and COMET on our validation set.

\section{Human Evaluation Methodology}\label{app:human-eval-methodology}

We use a customized version of the Appraise platform~\cite{federmann-2018-appraise}.
Raters are shown two system outputs per segment (for fluency) or per document (for accuracy).
Fluency is evaluated monolingually, i.e., raters do not see the source text, while accuracy is evaluated bilingually.
The systems are selected randomly per segment (for fluency) or per document (for accuracy) out of a set of four: the human reference from the WMT24++ benchmark~\cite{vamvas-et-al-2025-expanding}, Gemini 3 Pro, NLLB trained with the baseline LR$\rightarrow$HR augmentation approach, and NLLB trained with LR$\rightarrow$HR augmentation using dictionary prompting.
Screenshots of the evaluation interface are shown in Appendices~\ref{app:annotation-interface-fluency} and \ref{app:annotation-interface-accuracy}; the evaluation guidelines are provided in Appendix~\ref{app:eval-guidelines}.

In addition to the document-level accuracy ratings, we encouraged raters to indicate segment-level pairwise preferences, by selecting the better translation among the two systems.
The main human evaluation results are reported in Figure~\ref{fig:results-human}; segment-level pairwise win-rates are reported in Appendix~\ref{app:detailed-human-results}.
\clearpage

\onecolumn

\section{Datasets used for MT Training}\label{app:datasets}

\subsection{Parallel Datasets}

\begin{table}[H]
\centering
\footnotesize
\begin{tabularx}{\columnwidth}{@{}XXcrr@{}}
\toprule
\textbf{Name} & \textbf{Romansh varieties} & \textbf{URL} & \textbf{\textsc{rm} tokens} & \textbf{\textsc{de} tokens} \\
\midrule
Dictionary data & RG, Surs., Suts., Surm., Puter, Vall. & (\href{https://pledarigrond.ch/}{\ExternalLink}) & 3,298,051 & 6,530,231 \\
\midrule
Mediomatix \cite{hopton-et-al-2025-mediomatix} & Surs., Suts., Surm., Puter, Vall. & \href{https://huggingface.co/datasets/ZurichNLP/mediomatix}{\ExternalLink} & \dag{}\,6,946,947 & - \\
\midrule
Press releases of Canton Grisons & RG & (\href{https://www.gr.ch/RM/medias/communicaziuns/MMStaka/Seiten/AktuelleMeldungen.aspx}{\ExternalLink}) & 1,840,537 & 1,397,196 \\
\midrule
SwissLawTranslations \cite{niklaus-etal-2025-swiltra} & RG & \href{https://huggingface.co/datasets/joelniklaus/SwissLawTranslations}{\ExternalLink} & 607,517 & 494,980 \\
\midrule
Laws of Canton Grisons & RG & \href{https://huggingface.co/datasets/ZurichNLP/romansh-canton-laws}{\ExternalLink} & 538,174 & 402,047 \\
\midrule
Parallel data contributed by RTR & RG, Surs., Suts., Surm., Puter, Vall. & - & 646,856 & - \\
\midrule
Storyweaver & Surs., Suts., Surm., Puter, Vall. & - & 48,475 & 9,170 \\
\midrule
\textbf{Total} & & & 13,926,557 & 8,833,624 \\
\bottomrule
\end{tabularx}

\caption{Parallel datasets included in the MT training. Tokens are counted based on whitespace tokenization. \dag{}~Counting each pairwise combination of varieties separately.}
\label{tab:parallel-data}
\end{table}

\subsection{Monolingual Datasets}

\begin{table}[H]
\centering
\footnotesize
\begin{tabularx}{\columnwidth}{@{}Xcr@{}}
\toprule
\textbf{Name} & \textbf{URL} & \textbf{Tokens} \\
\midrule
Europarl \cite{koehn-2005-europarl} & \href{https://opus.nlpl.eu/synthetic/Europarl.php}{\ExternalLink} & (deduplicated) 52,239,860 \\
\midrule
Newscrawl 2023 \cite{kocmi-etal-2025-findings} & \href{https://www2.statmt.org/wmt25/mtdata/}{\ExternalLink} & $\sim$894,960,000 \\
\midrule
FineWeb2 \cite{penedo2025fineweb2pipelinescale} & \href{https://huggingface.co/datasets/HuggingFaceFW/fineweb-2}{\ExternalLink} & $\sim$262,271,052,000 \\
\bottomrule
\end{tabularx}

\caption{Monolingual German datasets used for the two \textbf{HR$\rightarrow$LR augmentation} variants. Token counts are approximate. Note that we sample a subset from these datasets, independently for each Romansh target variety.}
\label{tab:monolingual-german-data}
\end{table}

\begin{table}[H]
\centering
\footnotesize
\begin{tabularx}{\columnwidth}{@{}XXcr@{}}
\toprule
\textbf{Name} & \textbf{Romansh varieties} & \textbf{URL} & \textbf{Tokens} \\
\midrule
FineWeb2 \cite{penedo2025fineweb2pipelinescale} & RG & \href{https://huggingface.co/datasets/HuggingFaceFW/fineweb-2}{\ExternalLink} & 48,340,896 \\
\midrule
\mbox{La~Quotidiana} (1997–2008,~2021–2025) & RG, Surs., Suts., Surm., Puter, Vall. & \href{https://huggingface.co/datasets/ZurichNLP/quotidiana}{\ExternalLink} & 38,993,608 \\
\midrule
News articles by RTR & RG & - & 22,705,038 \\
\midrule
FinePDFs \cite{kydlicek2025finepdfs} & RG & \href{https://huggingface.co/datasets/HuggingFaceFW/finepdfs}{\ExternalLink} & 18,856,327 \\
\midrule
Mediomatix (unaligned) \cite{hopton-et-al-2025-mediomatix} & Surs., Suts., Surm., Puter, Vall. & \href{https://huggingface.co/datasets/ZurichNLP/mediomatix-raw}{\ExternalLink} & 4,639,563 \\
\midrule
FineWiki \cite{penedo2025finewiki} & RG & \href{https://huggingface.co/datasets/HuggingFaceFW/finewiki}{\ExternalLink} & 2,827,954 \\
\midrule
Audio transcriptions by RTR & RG, Surs., Suts., Surm., Puter, Vall. & (\href{https://developer.srgssr.ch/en/apis/rtr-linguistic}{\ExternalLink}) & 1,467,957 \\
\midrule
Theater plays & RG, Surs., Suts., Surm., Puter, Vall. & \href{https://huggingface.co/datasets/ZurichNLP/romansh_theater_plays}{\ExternalLink} & 1,079,943 \\
\midrule
Municipal documents & Surs., Suts., Surm., Vall. & \href{https://huggingface.co/datasets/ZurichNLP/romansh-municipal-text-corpus}{\ExternalLink} & 318,308 \\
\midrule
Historical Dictionary of Switzerland & RG & (\href{https://hls-dhs-dss.ch/rm/}{\ExternalLink}) & 234,715 \\
\midrule
Revista digl noss Sulom & Surm. & - & 187,247 \\
\midrule
Babulins & RG, Surs., Suts., Surm., Puter, Vall. & - & 49,444 \\
\midrule
\textbf{Total} & & & 139,701,000 \\
\bottomrule
\end{tabularx}

\caption{Monolingual Romansh datasets used for \textbf{LR$\rightarrow$HR augmentation}. Tokens are counted based on whitespace tokenization.}
\label{tab:monolingual-romansh-data}
\end{table}

\clearpage

\section{Statistics of LLM Requests}\label{app:llm-requests-stats}

\begin{table}[H]
\centering
\footnotesize
\begin{tabularx}{\columnwidth}{@{}Xrrrr@{}}
\toprule
 & \textbf{Requests} & \textbf{Total input tokens} & \textbf{Total output tokens} & \textbf{Approx. API cost} \\
\midrule
\multicolumn{5}{@{}l}{HR$\rightarrow$LR augmentation} \\
\quad\textendash~Europarl & 251,432 & 283,562,126 & 252,592,957 & 350 USD \\
\quad\textendash~Newscrawl / FineWeb2 & 2,615,424 & 1,163,299,838 & 267,593,106 & 500 USD \\
\midrule
\multicolumn{5}{@{}l}{LR$\rightarrow$HR augmentation} \\
\quad\textendash~Baseline prompt & 765,908 & 530,210,202 & 194,726,402 & 315 USD \\
\quad\textendash~Dictionary prompting & 765,908 & 8,292,343,080 & 196,047,179 & 1260 USD \\
\bottomrule
\end{tabularx}

\caption{Statistics of LLM requests for different data augmentation strategies. Tokens are counted in terms of the tokenizer of Gemini 2.5 Flash.
The variants required a different number of requests (and consequently, few-shot tokens) because the average document length can vary.
Furthermore, different tokenizer fertilities for German and Romansh lead to different input/output token ratios.}
\label{tab:llm-requests-stats}
\end{table}

\section{Statistics of Preprocessed Training Data}\label{app:training-data-statistics}

The following tables report training data statistics per language, before upsampling or reversing the translation directions for bidirectional training. Percentages in parentheses indicate the proportion of segments/tokens where either the sample itself or the corresponding source or target segment is synthetic. Tokens are counted based on whitespace tokenization.

\begin{table}[H]
\centering
\footnotesize
\begin{tabularx}{\columnwidth}{@{}Xcccc@{}}
\toprule
\textbf{Language} & \multicolumn{2}{c}{\textbf{Europarl}} & \multicolumn{2}{c}{\textbf{Newscrawl / FineWeb2}} \\
\cmidrule(lr){2-3} \cmidrule(lr){4-5}
 & \textbf{\# Segments} & \textbf{\# Tokens} & \textbf{\# Segments} & \textbf{\# Tokens} \\
\midrule
German & \phantom{0}7.8M \textcolor{gray!70}{(70\% synthetic)} & 121.0M \textcolor{gray!70}{(94\% synthetic)} & \phantom{0}\phantom{0}6.5M \textcolor{gray!70}{(64\% synthetic)} & 118.7M \textcolor{gray!70}{(94\% synthetic)} \\
RG & \phantom{0}2.7M \textcolor{gray!70}{(71\% synthetic)} & \phantom{0}49.8M \textcolor{gray!70}{(92\% synthetic)} & \phantom{0}\phantom{0}2.7M \textcolor{gray!70}{(70\% synthetic)} & \phantom{0}62.7M \textcolor{gray!70}{(94\% synthetic)} \\
Sursilvan & \phantom{0}0.7M \textcolor{gray!70}{(60\% synthetic)} & \phantom{0}11.1M \textcolor{gray!70}{(85\% synthetic)} & \phantom{0}\phantom{0}0.6M \textcolor{gray!70}{(55\% synthetic)} & \phantom{0}12.0M \textcolor{gray!70}{(86\% synthetic)} \\
Sutsilvan & \phantom{0}0.6M \textcolor{gray!70}{(25\% synthetic)} & \phantom{0}\phantom{0}5.0M \textcolor{gray!70}{(66\% synthetic)} & \phantom{0}\phantom{0}0.6M \textcolor{gray!70}{(24\% synthetic)} & \phantom{0}\phantom{0}6.1M \textcolor{gray!70}{(72\% synthetic)} \\
Surmiran & \phantom{0}1.4M \textcolor{gray!70}{(83\% synthetic)} & \phantom{0}26.8M \textcolor{gray!70}{(97\% synthetic)} & \phantom{0}\phantom{0}0.8M \textcolor{gray!70}{(70\% synthetic)} & \phantom{0}17.8M \textcolor{gray!70}{(96\% synthetic)} \\
Puter & \phantom{0}0.8M \textcolor{gray!70}{(40\% synthetic)} & \phantom{0}\phantom{0}9.3M \textcolor{gray!70}{(81\% synthetic)} & \phantom{0}\phantom{0}0.8M \textcolor{gray!70}{(42\% synthetic)} & \phantom{0}13.0M \textcolor{gray!70}{(86\% synthetic)} \\
Vallader & \phantom{0}2.0M \textcolor{gray!70}{(74\% synthetic)} & \phantom{0}36.5M \textcolor{gray!70}{(95\% synthetic)} & \phantom{0}\phantom{0}1.4M \textcolor{gray!70}{(63\% synthetic)} & \phantom{0}31.2M \textcolor{gray!70}{(94\% synthetic)} \\
\midrule
\textbf{Total} & 16.1M \textcolor{gray!70}{(68\% synthetic)} & 259.5M \textcolor{gray!70}{(93\% synthetic)} & \phantom{0}13.4M \textcolor{gray!70}{(62\% synthetic)} & 261.4M \textcolor{gray!70}{(93\% synthetic)} \\
\bottomrule
\end{tabularx}
\caption{Training data statistics for the HR$\rightarrow$LR direction.}
\label{tab:training-stats-hr-to-lr}
\end{table}

\begin{table}[H]
\centering
\footnotesize
\begin{tabularx}{\columnwidth}{@{}Xcccc@{}}
\toprule
\textbf{Language} & \multicolumn{2}{c}{\textbf{Baseline prompt}} & \multicolumn{2}{c}{\textbf{Dictionary prompting}} \\
\cmidrule(lr){2-3} \cmidrule(lr){4-5}
 & \textbf{\# Segments} & \textbf{\# Tokens} & \textbf{\# Segments} & \textbf{\# Tokens} \\
\midrule
German & \phantom{0}6.3M \textcolor{gray!70}{(63\% synthetic)} & 115.3M \textcolor{gray!70}{(94\% synthetic)} & \phantom{0}6.4M \textcolor{gray!70}{(64\% synthetic)} & 117.5M \textcolor{gray!70}{(94\% synthetic)} \\
RG & \phantom{0}3.3M \textcolor{gray!70}{(76\% synthetic)} & \phantom{0}67.1M \textcolor{gray!70}{(94\% synthetic)} & \phantom{0}3.4M \textcolor{gray!70}{(76\% synthetic)} & \phantom{0}68.9M \textcolor{gray!70}{(94\% synthetic)} \\
Sursilvan & \phantom{0}1.1M \textcolor{gray!70}{(74\% synthetic)} & 33.1M \textcolor{gray!70}{(95\% synthetic)} & \phantom{0}1.1M \textcolor{gray!70}{(75\% synthetic)} & \phantom{0}33.9M \textcolor{gray!70}{(95\% synthetic)} \\
Sutsilvan & \phantom{0}0.5M \textcolor{gray!70}{(14\% synthetic)} & \phantom{0}\phantom{0}4.4M \textcolor{gray!70}{(61\% synthetic)} & \phantom{0}0.5M \textcolor{gray!70}{(14\% synthetic)} & \phantom{0}\phantom{0}4.5M \textcolor{gray!70}{(61\% synthetic)} \\
Surmiran & \phantom{0}0.4M \textcolor{gray!70}{(37\% synthetic)} & \phantom{0}\phantom{0}6.7M \textcolor{gray!70}{(89\% synthetic)} & \phantom{0}0.4M \textcolor{gray!70}{(38\% synthetic)} & \phantom{0}\phantom{0}6.8M \textcolor{gray!70}{(89\% synthetic)} \\
Puter & \phantom{0}0.7M \textcolor{gray!70}{(25\% synthetic)} & \phantom{0}\phantom{0}6.2M \textcolor{gray!70}{(71\% synthetic)} & \phantom{0}0.7M \textcolor{gray!70}{(25\% synthetic)} & \phantom{0}\phantom{0}6.3M \textcolor{gray!70}{(71\% synthetic)} \\
Vallader & \phantom{0}0.8M \textcolor{gray!70}{(34\% synthetic)} & \phantom{0}13.3M \textcolor{gray!70}{(86\% synthetic)} & \phantom{0}0.8M \textcolor{gray!70}{(35\% synthetic)} & \phantom{0}13.5M \textcolor{gray!70}{(86\% synthetic)} \\
\midrule
\textbf{Total} & 13.1M \textcolor{gray!70}{(61\% synthetic)} & 246.1M \textcolor{gray!70}{(92\% synthetic)} & 13.3M \textcolor{gray!70}{(62\% synthetic)} & 251.3M \textcolor{gray!70}{(93\% synthetic)} \\
\bottomrule
\end{tabularx}

\caption{Training data statistics for the LR$\rightarrow$HR direction.}
\label{tab:training-stats-lr-to-hr}
\end{table}

\clearpage

\section{Automatic Evaluation: Romansh$\rightarrow$German BLEU Scores}\label{app:detailed-rm-to-de-bleu}

\begin{table}[H]
\centering

{\footnotesize
\begin{tabular*}{\textwidth}{@{\extracolsep{\fill}}lrrrrrrcr@{}}
\toprule
\mbox{\textbf{System}} & \multicolumn{6}{c}{\textbf{WMT24++}} & \textbf{BOUQuET} & \multirow{2}{*}{\textbf{Avg.}} \\
\cmidrule(lr){2-7} \cmidrule(lr){8-8} \cmidrule(lr){9-9}
 & \textbf{RG} & \textbf{Sursilvan} & \textbf{Sutsilvan} & \textbf{Surmiran} & \textbf{Puter} & \textbf{Vallader} & \textbf{RG} \\
\midrule
\mbox{Gemini 2.5 Flash} & 55.7 & 50.0 & 44.9 & 49.9 & 52.0 & 58.9 & \textbf{33.9} & 49.3 \\
\mbox{Gemini 3 Pro (preview)} & \textbf{55.8} & \textbf{50.8} & \textbf{48.7} & 52.2 & \textbf{53.8} & \textbf{60.8} & 33.8 & \textbf{50.8} \\
\mbox{Gemini 3.7 Flash} & 55.2 & 50.3 & 48.2 & \textbf{52.9} & 53.3 & 60.6 & 33.6 & 50.6 \\
\midrule
\mbox{\textit{Fine-tuned NLLB}} & & & & & & & & \\[0.35em]
\mbox{No data augmentation} & 38.3 & 33.9 & 33.3 & 33.7 & 34.6 & 37.2 & 26.5 & 33.9 \\[0.45em]
\mbox{HR$\rightarrow$LR augmentation:} & & & & & & & & \\[0.32em]
\mbox{– Europarl data} & 45.8 & 42.2 & 40.9 & 43.9 & 45.5 & 49.9 & 30.1 & 42.6 \\
\mbox{– Newscrawl / FineWeb2} & \underline{50.5} & 46.1 & \underline{44.5} & \underline{47.4} & \underline{49.6} & 54.7 & 33.1 & \underline{46.6} \\[0.45em]
\mbox{LR$\rightarrow$HR augmentation:} & & & & & & & & \\[0.32em]
\mbox{– Baseline prompt} & 50.1 & 46.0 & 43.9 & 46.8 & 49.3 & \underline{54.8} & 33.0 & 46.3 \\
\mbox{– Dictionary prompting} & \underline{50.5} & \underline{46.3} & 44.2 & 47.3 & 49.5 & \underline{54.8} & \underline{33.3} & 46.5 \\
\bottomrule
\end{tabular*}
}

\caption{Per-variety BLEU scores for Romansh$\rightarrow$German translation on the second half of the WMT24++ benchmark, together with BOUQuET results for Rumantsch Grischun. \textbf{Bold} denotes the best system; \underline{underline} indicates the best variant among the fine-tuned NLLB models; the last column reports the arithmetic mean of the seven columns.}
\label{tab:detailed-rm-to-de-bleu}
\end{table}

\vfill

\section{Prompt Templates}\label{sec:prompt-templates}

\subsection{Baseline prompt}\label{app:forward-backtranslation}

\begin{quote}
\ttfamily
Translate the following text into \{target\_language\}. The text is written in \{source\_language\} and is provided to you wrapped in triple backticks (\textasciigrave\textasciigrave\textasciigrave). Just answer with the translation and nothing else.\textbackslash n\textbackslash n\textasciigrave\textasciigrave\textasciigrave\{source\_document\}\textasciigrave\textasciigrave\textasciigrave\textbackslash n
\end{quote}

\subsection{Prompt with Dictionary Information}\label{app:backtranslation-dict}

\begin{quote}
\ttfamily
Translate the text below into \{target\_language\}. The text is written in \{source\_language\} and is provided to you wrapped in triple backticks (\textasciigrave\textasciigrave\textasciigrave).\textbackslash n\textbackslash nSome of the dictionary entries below might be helpful for translating the text:\textbackslash n\textbackslash n\{dictionary\_entries\}\textbackslash n\textbackslash nJust answer with the translation and nothing else.\textbackslash n\textbackslash nText to translate:\textbackslash n\textasciigrave\textasciigrave\textasciigrave\{source\_segment\}\textasciigrave\textasciigrave\textasciigrave
\end{quote}

\vfill

\clearpage

\section{Full Example for a Prompt with Dictionary Information}\label{app:dictionary-prompt-example}
\begingroup
\footnotesize\ttfamily
\texttt{Translate the text below into German. The text is written in Romansh (Sursilvan variety) and is provided to you wrapped in triple backticks (\textasciigrave\textasciigrave\textasciigrave).\\\\Some of the dictionary entries below might be helpful for translating the text:\\\\- stos (lemma: stuer) → müssen [PoS=V;Mood=IND;Tense=PRS;Person=2;Number=SG]\\- violenza (lemma: violenza) → Heftigkeit [PoS=N;Gender=FEM;Number=SG]\\- violenza (lemma: violenza) → Gewalt [PoS=N;Gender=FEM;Number=SG]\\- violenza (lemma: violenza) → Stärke [PoS=N;Gender=FEM;Number=SG]\\- violenza (lemma: violenza) → Wucht [PoS=N;Gender=FEM;Number=SG]\\- violenza (lemma: violenza) → Gewalttätigkeit [PoS=N;Gender=FEM;Number=SG]\\- violenza (lemma: violenza) → Tätlichkeit [PoS=N;Gender=FEM;Number=SG]\\- violenza (lemma: violenza) → Vergewaltigung [PoS=N;Gender=FEM;Number=SG]\\- violenza (lemma: violenza) → Notzucht [PoS=N;Gender=FEM;Number=SG]\\- meina (lemma: meina) → führen [PoS=V;Finiteness=NFIN]\\- meina (lemma: meina) → ausführen [PoS=V;Finiteness=NFIN]\\- meina (lemma: meina) → transportieren [PoS=V;Finiteness=NFIN]\\- meina (lemma: meina) → leiten [PoS=V;Finiteness=NFIN]\\- meina (lemma: meina) → anführen [PoS=V;Finiteness=NFIN]\\- meina (lemma: meina) → vorstehen [PoS=V;Finiteness=NFIN]\\- meina (lemma: meina) → regieren [PoS=V;Finiteness=NFIN]\\- meina (lemma: meina) → verbringen [PoS=V;Finiteness=NFIN]\\- meina (lemma: meina) → durchführen [PoS=V;Finiteness=NFIN]\\- meina (lemma: meina) → vollenden [PoS=V;Finiteness=NFIN]\\- meina (lemma: meina) → verwalten [PoS=V;Finiteness=NFIN]\\- meina (lemma: meina) → fahren [PoS=V;Finiteness=NFIN]\\- meina (lemma: meina) → steuern [PoS=V;Finiteness=NFIN]\\- meina (lemma: meina) → lenken [PoS=V;Finiteness=NFIN]\\- meina (lemma: meina) → wenden [PoS=V;Finiteness=NFIN]\\- meina (lemma: meina) → drehen [PoS=V;Finiteness=NFIN]\\- meina (lemma: meina) → kehren [PoS=V;Finiteness=NFIN]\\- meina (lemma: meina) → richten [PoS=V;Finiteness=NFIN]\\- meina (lemma: meina) → decken [PoS=V;Finiteness=NFIN]\\- violenza (lemma: violenza) → Heftigkeit [PoS=N;Gender=FEM;Number=SG]\\- violenza (lemma: violenza) → Gewalt [PoS=N;Gender=FEM;Number=SG]\\- violenza (lemma: violenza) → Stärke [PoS=N;Gender=FEM;Number=SG]\\- violenza (lemma: violenza) → Wucht [PoS=N;Gender=FEM;Number=SG]\\- violenza (lemma: violenza) → Gewalttätigkeit [PoS=N;Gender=FEM;Number=SG]\\- violenza (lemma: violenza) → Tätlichkeit [PoS=N;Gender=FEM;Number=SG]\\- violenza (lemma: violenza) → Vergewaltigung [PoS=N;Gender=FEM;Number=SG]\\- violenza (lemma: violenza) → Notzucht [PoS=N;Gender=FEM;Number=SG]\\- decisiun (lemma: decisiun) → Entscheid [PoS=N;Gender=FEM;Number=SG]\\- decisiun (lemma: decisiun) → Entscheidung [PoS=N;Gender=FEM;Number=SG]\\- decisiun (lemma: decisiun) → Abmachung [PoS=N;Gender=FEM;Number=SG]\\- decisiun (lemma: decisiun) → Entschluss [PoS=N;Gender=FEM;Number=SG]\\- decisiun (lemma: decisiun) → Entschliessung [PoS=N;Gender=FEM;Number=SG]\\- dedicau (lemma: dedicar) → widmen [PoS=V;VerbForm=PTCP;Tense=PST;Gender=NEUT]\\- protecziun (lemma: protecziun) → Schutz [PoS=N;Gender=FEM;Number=SG]\\- protecziun (lemma: protecziun) → Beschützung [PoS=N;Gender=FEM;Number=SG]\\- protecziun (lemma: protecziun) → Bewahrung [PoS=N;Gender=FEM;Number=SG]\\- protecziun (lemma: protecziun) → Wehr [PoS=N;Gender=FEM;Number=SG]\\- protecziun (lemma: protecziun) → Protektion [PoS=N;Gender=FEM;Number=SG]\\- protecziun (lemma: protecziun) → Begünstigung [PoS=N;Gender=FEM;Number=SG]\\- protecziun (lemma: protecziun) → Förderung [PoS=N;Gender=FEM;Number=SG]\\- protecziun (lemma: protecziun) → Gönnerschaft [PoS=N;Gender=FEM;Number=SG]\\\\Just answer with the translation and nothing else.\\\\Text to translate:\\\textasciigrave\textasciigrave\textasciigrave Quei hai jeu buca detg. Quei stos ti era buc. Ti sas meglier che tut ils auters che violenza meina mintga ga a dapli violenza. Il cussegl ha priu la decisiun. Il cussegl dat ei buca pli! Nus havein dedicau noss'entira veta, mellis onns, alla protecziun dalla veta.\textasciigrave\textasciigrave\textasciigrave }
\par
\endgroup

\vfill
\clearpage

\section{Generative Confusion Matrices}\label{app:confusion-matrices-all-systems}

\begin{figure}[H]
\centering
\begin{subfigure}[b]{0.48\textwidth}
\centering

\begin{tabularx}{8cm}{@{}p{0.9cm}|*{6}{>{\raggedleft\arraybackslash}p{0.65cm}}}
\multirow[b]{2}{*}{$\downarrow$~\textbf{tgt}}
& \multicolumn{6}{l}{$\textbf{ref}$~$\rightarrow$} \\
& RG & Surs.\ & Suts.\ & Surm.\ & Puter & Vall. \\
RG     & \cellcolor{uzhblue!40}40.3   & \cellcolor{uzhblue!14}18.4   & \cellcolor{uzhblue!3}8.9   & \cellcolor{uzhblue!7}11.8   & \cellcolor{uzhblue!7}12.2   & \cellcolor{uzhblue!11}15.4   \\
Surs.  & \cellcolor{uzhblue!25}27.2 & \cellcolor{uzhblue!26}28.0 & \cellcolor{uzhblue!3}8.6 & \cellcolor{uzhblue!5}10.5 & \cellcolor{uzhblue!5}10.0 & \cellcolor{uzhblue!8}12.7 \\
Suts.  & \cellcolor{uzhblue!19}22.0 & \cellcolor{uzhblue!12}16.4 & \cellcolor{uzhblue!5}10.6 & \cellcolor{uzhblue!8}12.6 & \cellcolor{uzhblue!5}10.0 & \cellcolor{uzhblue!7}12.1 \\
Surm.  & \cellcolor{uzhblue!18}21.6 & \cellcolor{uzhblue!9}13.7 & \cellcolor{uzhblue!4}9.4 & \cellcolor{uzhblue!13}16.9 & \cellcolor{uzhblue!6}11.2 & \cellcolor{uzhblue!8}12.5 \\
Puter  & \cellcolor{uzhblue!10}14.8   & \cellcolor{uzhblue!5}10.6   & \cellcolor{uzhblue!2}7.5   & \cellcolor{uzhblue!3}8.6   & \cellcolor{uzhblue!19}22.0   & \cellcolor{uzhblue!22}24.6   \\
Vall.  & \cellcolor{uzhblue!12}16.3 & \cellcolor{uzhblue!6}11.1 & \cellcolor{uzhblue!2}7.6 & \cellcolor{uzhblue!3}8.7 & \cellcolor{uzhblue!17}20.6 & \cellcolor{uzhblue!25}27.3 \\
\end{tabularx}

\caption{\mbox{Gemini 2.5 Flash}}
\label{subfig:confusion_gemini_25_flash}
\end{subfigure}\hfill\begin{subfigure}[b]{0.48\textwidth}
\centering

\begin{tabularx}{6cm}{@{}*{6}{>{\raggedleft\arraybackslash}p{0.65cm}}}
\multicolumn{6}{l}{$\textbf{ref}$~$\rightarrow$} \\
RG & Surs.\ & Suts.\ & Surm.\ & Puter & Vall. \\
\cellcolor{uzhblue!46}45.3   & \cellcolor{uzhblue!16}19.8   & \cellcolor{uzhblue!4}9.1   & \cellcolor{uzhblue!8}12.8   & \cellcolor{uzhblue!8}12.6   & \cellcolor{uzhblue!12}16.2   \\
\cellcolor{uzhblue!18}21.7 & \cellcolor{uzhblue!37}37.1 & \cellcolor{uzhblue!3}9.0 & \cellcolor{uzhblue!5}10.5 & \cellcolor{uzhblue!4}9.2 & \cellcolor{uzhblue!6}11.1 \\
\cellcolor{uzhblue!12}16.4 & \cellcolor{uzhblue!10}14.3 & \cellcolor{uzhblue!13}17.3 & \cellcolor{uzhblue!10}14.9 & \cellcolor{uzhblue!5}10.0 & \cellcolor{uzhblue!6}11.1 \\
\cellcolor{uzhblue!12}16.5 & \cellcolor{uzhblue!8}12.5 & \cellcolor{uzhblue!5}10.2 & \cellcolor{uzhblue!25}27.1 & \cellcolor{uzhblue!5}10.0 & \cellcolor{uzhblue!5}10.7 \\
\cellcolor{uzhblue!9}13.6   & \cellcolor{uzhblue!4}9.8   & \cellcolor{uzhblue!1}7.1   & \cellcolor{uzhblue!4}9.2   & \cellcolor{uzhblue!34}34.7   & \cellcolor{uzhblue!21}24.1   \\
\cellcolor{uzhblue!13}17.3 & \cellcolor{uzhblue!6}11.5 & \cellcolor{uzhblue!2}7.9 & \cellcolor{uzhblue!4}9.7 & \cellcolor{uzhblue!20}23.2 & \cellcolor{uzhblue!35}36.1 \\
\end{tabularx}

\caption{\mbox{Gemini 3 Pro}}
\label{subfig:confusion_gemini_3_pro}
\end{subfigure}
\vspace{0.5cm}

\begin{subfigure}[b]{0.48\textwidth}
\centering

\begin{tabularx}{8cm}{@{}p{0.9cm}|*{6}{>{\raggedleft\arraybackslash}p{0.65cm}}}
\multirow[b]{2}{*}{$\downarrow$~\textbf{tgt}}
& \multicolumn{6}{l}{$\textbf{ref}$~$\rightarrow$} \\
& RG & Surs.\ & Suts.\ & Surm.\ & Puter & Vall. \\
RG     & \cellcolor{uzhblue!41}40.9   & \cellcolor{uzhblue!14}18.1   & \cellcolor{uzhblue!3}8.6   & \cellcolor{uzhblue!6}11.5   & \cellcolor{uzhblue!7}12.0   & \cellcolor{uzhblue!10}15.0   \\
Surs.  & \cellcolor{uzhblue!19}22.5 & \cellcolor{uzhblue!33}34.2 & \cellcolor{uzhblue!3}8.8 & \cellcolor{uzhblue!5}10.4 & \cellcolor{uzhblue!4}9.5 & \cellcolor{uzhblue!6}11.6 \\
Suts.  & \cellcolor{uzhblue!12}16.3 & \cellcolor{uzhblue!10}14.5 & \cellcolor{uzhblue!10}14.8 & \cellcolor{uzhblue!9}13.5 & \cellcolor{uzhblue!4}9.6 & \cellcolor{uzhblue!6}10.9 \\
Surm.  & \cellcolor{uzhblue!12}16.7 & \cellcolor{uzhblue!7}12.0 & \cellcolor{uzhblue!5}10.0 & \cellcolor{uzhblue!19}22.1 & \cellcolor{uzhblue!5}10.7 & \cellcolor{uzhblue!6}11.2 \\
Puter  & \cellcolor{uzhblue!9}13.9   & \cellcolor{uzhblue!5}10.3   & \cellcolor{uzhblue!1}7.4   & \cellcolor{uzhblue!3}8.8   & \cellcolor{uzhblue!29}30.4   & \cellcolor{uzhblue!19}22.1   \\
Vall.  & \cellcolor{uzhblue!11}15.3 & \cellcolor{uzhblue!5}10.7 & \cellcolor{uzhblue!1}7.3 & \cellcolor{uzhblue!3}8.5 & \cellcolor{uzhblue!16}19.8 & \cellcolor{uzhblue!28}29.8 \\
\end{tabularx}

\caption{\mbox{Gemini 3.7 Flash}}
\label{subfig:confusion_gemini_37_flash}
\end{subfigure}\hfill\begin{subfigure}[b]{0.48\textwidth}
\centering

\begin{tabularx}{6cm}{@{}*{6}{>{\raggedleft\arraybackslash}p{0.65cm}}}
\multicolumn{6}{l}{$\textbf{ref}$~$\rightarrow$} \\
RG & Surs.\ & Suts.\ & Surm.\ & Puter & Vall. \\
\cellcolor{uzhblue!35}35.9   & \cellcolor{uzhblue!12}16.5   & \cellcolor{uzhblue!2}7.4   & \cellcolor{uzhblue!5}10.4   & \cellcolor{uzhblue!6}10.8   & \cellcolor{uzhblue!8}13.3   \\
\cellcolor{uzhblue!20}23.1 & \cellcolor{uzhblue!28}29.4 & \cellcolor{uzhblue!1}7.4 & \cellcolor{uzhblue!3}8.6 & \cellcolor{uzhblue!3}8.6 & \cellcolor{uzhblue!5}10.1 \\
\cellcolor{uzhblue!6}11.1 & \cellcolor{uzhblue!3}9.0 & \cellcolor{uzhblue!27}29.2 & \cellcolor{uzhblue!3}8.9 & \cellcolor{uzhblue!0}6.5 & \cellcolor{uzhblue!1}7.4 \\
\cellcolor{uzhblue!12}16.6 & \cellcolor{uzhblue!5}10.8 & \cellcolor{uzhblue!3}8.6 & \cellcolor{uzhblue!23}25.3 & \cellcolor{uzhblue!2}8.2 & \cellcolor{uzhblue!4}9.1 \\
\cellcolor{uzhblue!8}12.9   & \cellcolor{uzhblue!3}8.8   & \cellcolor{uzhblue!0}6.2   & \cellcolor{uzhblue!2}7.4   & \cellcolor{uzhblue!29}31.0   & \cellcolor{uzhblue!15}19.2   \\
\cellcolor{uzhblue!16}19.7 & \cellcolor{uzhblue!6}11.6 & \cellcolor{uzhblue!1}6.6 & \cellcolor{uzhblue!2}8.2 & \cellcolor{uzhblue!13}16.9 & \cellcolor{uzhblue!23}26.0 \\
\end{tabularx}

\caption{\mbox{No data augmentation}}
\label{subfig:confusion_no_data_aug}
\end{subfigure}
\vspace{0.5cm}

\begin{subfigure}[b]{0.48\textwidth}
\centering

\begin{tabularx}{8cm}{@{}p{0.9cm}|*{6}{>{\raggedleft\arraybackslash}p{0.65cm}}}
\multirow[b]{2}{*}{$\downarrow$~\textbf{tgt}}
& \multicolumn{6}{l}{$\textbf{ref}$~$\rightarrow$} \\
& RG & Surs.\ & Suts.\ & Surm.\ & Puter & Vall. \\
RG     & \cellcolor{uzhblue!41}40.8   & \cellcolor{uzhblue!13}17.3   & \cellcolor{uzhblue!2}7.8   & \cellcolor{uzhblue!6}11.1   & \cellcolor{uzhblue!6}11.4   & \cellcolor{uzhblue!10}14.7   \\
Surs.  & \cellcolor{uzhblue!21}23.6 & \cellcolor{uzhblue!30}31.3 & \cellcolor{uzhblue!2}7.7 & \cellcolor{uzhblue!4}9.4 & \cellcolor{uzhblue!3}8.7 & \cellcolor{uzhblue!6}11.1 \\
Suts.  & \cellcolor{uzhblue!18}21.6 & \cellcolor{uzhblue!10}14.8 & \cellcolor{uzhblue!7}12.4 & \cellcolor{uzhblue!8}13.0 & \cellcolor{uzhblue!4}9.8 & \cellcolor{uzhblue!7}12.0 \\
Surm.  & \cellcolor{uzhblue!19}22.0 & \cellcolor{uzhblue!8}12.7 & \cellcolor{uzhblue!4}9.4 & \cellcolor{uzhblue!15}18.7 & \cellcolor{uzhblue!6}10.9 & \cellcolor{uzhblue!7}12.2 \\
Puter  & \cellcolor{uzhblue!12}16.4   & \cellcolor{uzhblue!5}10.5   & \cellcolor{uzhblue!1}6.7   & \cellcolor{uzhblue!2}8.2   & \cellcolor{uzhblue!21}23.8   & \cellcolor{uzhblue!25}27.4   \\
Vall.  & \cellcolor{uzhblue!13}17.0 & \cellcolor{uzhblue!5}10.6 & \cellcolor{uzhblue!1}6.8 & \cellcolor{uzhblue!2}8.2 & \cellcolor{uzhblue!17}20.9 & \cellcolor{uzhblue!29}30.7 \\
\end{tabularx}

\caption{\mbox{HR$\rightarrow$LR augmentation with Europarl data}}
\label{subfig:confusion_forward_translation_europarl}
\end{subfigure}\hfill\begin{subfigure}[b]{0.48\textwidth}
\centering

\begin{tabularx}{6cm}{@{}*{6}{>{\raggedleft\arraybackslash}p{0.65cm}}}
\multicolumn{6}{l}{$\textbf{ref}$~$\rightarrow$} \\
RG & Surs.\ & Suts.\ & Surm.\ & Puter & Vall. \\
\cellcolor{uzhblue!42}41.4   & \cellcolor{uzhblue!14}17.9   & \cellcolor{uzhblue!2}8.2   & \cellcolor{uzhblue!6}11.2   & \cellcolor{uzhblue!6}11.6   & \cellcolor{uzhblue!11}15.1   \\
\cellcolor{uzhblue!21}24.0 & \cellcolor{uzhblue!29}30.4 & \cellcolor{uzhblue!2}7.9 & \cellcolor{uzhblue!4}9.8 & \cellcolor{uzhblue!4}9.3 & \cellcolor{uzhblue!6}11.4 \\
\cellcolor{uzhblue!16}19.6 & \cellcolor{uzhblue!11}15.5 & \cellcolor{uzhblue!7}12.1 & \cellcolor{uzhblue!8}13.2 & \cellcolor{uzhblue!4}9.7 & \cellcolor{uzhblue!7}11.7 \\
\cellcolor{uzhblue!17}20.5 & \cellcolor{uzhblue!8}13.2 & \cellcolor{uzhblue!4}9.2 & \cellcolor{uzhblue!15}18.5 & \cellcolor{uzhblue!5}10.6 & \cellcolor{uzhblue!7}11.7 \\
\cellcolor{uzhblue!11}15.1   & \cellcolor{uzhblue!5}10.1   & \cellcolor{uzhblue!1}6.9   & \cellcolor{uzhblue!3}8.4   & \cellcolor{uzhblue!22}24.6   & \cellcolor{uzhblue!23}25.4   \\
\cellcolor{uzhblue!12}16.5 & \cellcolor{uzhblue!5}10.6 & \cellcolor{uzhblue!1}7.0 & \cellcolor{uzhblue!2}8.2 & \cellcolor{uzhblue!18}21.0 & \cellcolor{uzhblue!28}29.4 \\
\end{tabularx}

\caption{\mbox{HR$\rightarrow$LR augmentation with Newscrawl / FineWeb2 data}}
\label{subfig:confusion_forward_translation_newscrawl_fineweb2}
\end{subfigure}
\vspace{0.5cm}

\begin{subfigure}[b]{0.48\textwidth}
\centering

\begin{tabularx}{8cm}{@{}p{0.9cm}|*{6}{>{\raggedleft\arraybackslash}p{0.65cm}}}
\multirow[b]{2}{*}{$\downarrow$~\textbf{tgt}}
& \multicolumn{6}{l}{$\textbf{ref}$~$\rightarrow$} \\
& RG & Surs.\ & Suts.\ & Surm.\ & Puter & Vall. \\
RG     & \cellcolor{uzhblue!50}48.1   & \cellcolor{uzhblue!16}19.9   & \cellcolor{uzhblue!3}8.5   & \cellcolor{uzhblue!7}12.2   & \cellcolor{uzhblue!7}12.4   & \cellcolor{uzhblue!11}15.7   \\
Surs.  & \cellcolor{uzhblue!15}18.9 & \cellcolor{uzhblue!45}44.3 & \cellcolor{uzhblue!2}8.3 & \cellcolor{uzhblue!4}9.4 & \cellcolor{uzhblue!2}8.2 & \cellcolor{uzhblue!4}9.9 \\
Suts.  & \cellcolor{uzhblue!3}8.9 & \cellcolor{uzhblue!4}9.2 & \cellcolor{uzhblue!41}40.6 & \cellcolor{uzhblue!4}9.9 & \cellcolor{uzhblue!1}6.7 & \cellcolor{uzhblue!2}7.4 \\
Surm.  & \cellcolor{uzhblue!8}12.7 & \cellcolor{uzhblue!5}10.5 & \cellcolor{uzhblue!4}9.7 & \cellcolor{uzhblue!43}42.3 & \cellcolor{uzhblue!2}8.2 & \cellcolor{uzhblue!3}8.9 \\
Puter  & \cellcolor{uzhblue!7}11.8   & \cellcolor{uzhblue!3}8.8   & \cellcolor{uzhblue!0}6.3   & \cellcolor{uzhblue!2}8.0   & \cellcolor{uzhblue!46}44.8   & \cellcolor{uzhblue!22}24.8   \\
Vall.  & \cellcolor{uzhblue!10}14.6 & \cellcolor{uzhblue!4}9.9 & \cellcolor{uzhblue!1}6.7 & \cellcolor{uzhblue!2}8.0 & \cellcolor{uzhblue!22}25.0 & \cellcolor{uzhblue!45}44.3 \\
\end{tabularx}

\caption{\mbox{LR$\rightarrow$HR augmentation}}
\label{subfig:confusion_back_translation}
\end{subfigure}\hfill\begin{subfigure}[b]{0.48\textwidth}
\centering

\begin{tabularx}{6cm}{@{}*{6}{>{\raggedleft\arraybackslash}p{0.65cm}}}
\multicolumn{6}{l}{$\textbf{ref}$~$\rightarrow$} \\
RG & Surs.\ & Suts.\ & Surm.\ & Puter & Vall. \\
\cellcolor{uzhblue!50}48.4   & \cellcolor{uzhblue!16}19.9   & \cellcolor{uzhblue!3}8.6   & \cellcolor{uzhblue!7}12.2   & \cellcolor{uzhblue!7}12.4   & \cellcolor{uzhblue!11}15.5   \\
\cellcolor{uzhblue!15}19.0 & \cellcolor{uzhblue!45}44.5 & \cellcolor{uzhblue!3}8.3 & \cellcolor{uzhblue!4}9.4 & \cellcolor{uzhblue!2}8.2 & \cellcolor{uzhblue!4}9.8 \\
\cellcolor{uzhblue!3}9.0 & \cellcolor{uzhblue!4}9.3 & \cellcolor{uzhblue!41}40.5 & \cellcolor{uzhblue!5}10.1 & \cellcolor{uzhblue!1}6.7 & \cellcolor{uzhblue!2}7.5 \\
\cellcolor{uzhblue!7}12.5 & \cellcolor{uzhblue!5}10.5 & \cellcolor{uzhblue!4}9.7 & \cellcolor{uzhblue!44}43.0 & \cellcolor{uzhblue!2}8.1 & \cellcolor{uzhblue!3}8.9 \\
\cellcolor{uzhblue!7}11.9   & \cellcolor{uzhblue!3}8.9   & \cellcolor{uzhblue!0}6.3   & \cellcolor{uzhblue!2}8.0   & \cellcolor{uzhblue!46}44.9   & \cellcolor{uzhblue!22}25.0   \\
\cellcolor{uzhblue!10}14.8 & \cellcolor{uzhblue!5}10.1 & \cellcolor{uzhblue!1}6.7 & \cellcolor{uzhblue!2}8.1 & \cellcolor{uzhblue!22}24.7 & \cellcolor{uzhblue!46}44.6 \\
\end{tabularx}

\caption{\mbox{LR$\rightarrow$HR augmentation with Dictionary prompting}}
\label{subfig:confusion_dict_prompting}
\end{subfigure}
\caption{Confusion matrices similar to Figure~\ref{fig:confusion-matrices-main} illustrating the target variety adherence in German$\rightarrow$Romansh translation. Results are based on BLEU.}
\label{fig:confusion_matrices_all_systems}
\end{figure}

\vfill
\clearpage

\section{Detailed Human Evaluation Results}\label{app:detailed-human-results}

\subsection{Fluency}

\begin{table}[H]
\centering

\footnotesize
\begin{tabularx}{\columnwidth}{@{}Xrrrrrr@{}}
\toprule
\textbf{System} & \textbf{RG} & \textbf{Surs.} & \textbf{Suts.} & \textbf{Surm.} & \textbf{Puter} & \textbf{Vall.} \\
\midrule
Human reference & 0.24 $\pm$ 0.10 & 0.66 $\pm$ 0.07 & 0.62 $\pm$ 0.07 & 0.60 $\pm$ 0.08 & 0.50 $\pm$ 0.07 & 0.23 $\pm$ 0.12 \\
\midrule
Gemini 3 Pro (preview) & -0.30 $\pm$ 0.14 & -0.75 $\pm$ 0.11 & -1.35 $\pm$ 0.08 & -0.98 $\pm$ 0.11 & -0.92 $\pm$ 0.15 & -0.65 $\pm$ 0.13 \\
\mbox{LR$\rightarrow$HR augmented NLLB} & 0.03 $\pm$ 0.11 & 0.02 $\pm$ 0.12 & 0.36 $\pm$ 0.09 & 0.17 $\pm$ 0.10 & 0.05 $\pm$ 0.10 & 0.19 $\pm$ 0.12 \\
\quad + Dictionary prompting & 0.04 $\pm$ 0.11 & 0.10 $\pm$ 0.12 & 0.26 $\pm$ 0.09 & 0.28 $\pm$ 0.11 & 0.15 $\pm$ 0.12 & 0.20 $\pm$ 0.14 \\
\bottomrule
\end{tabularx} 
\normalsize

\caption{Per-variety fluency scores for German$\rightarrow$Romansh translation. Ratings were made on the segment level and then averaged across segments.}
\label{tab:detailed-human-fluency}
\end{table}

\subsection{Accuracy}

\begin{table}[H]
\centering

\footnotesize
\begin{tabularx}{\columnwidth}{@{}Xrrrrrr@{}}
\toprule
\textbf{System} & \textbf{RG} & \textbf{Surs.} & \textbf{Suts.} & \textbf{Surm.} & \textbf{Puter} & \textbf{Vall.} \\
\midrule
Human reference & 0.40 $\pm$ 0.27 & 0.74 $\pm$ 0.25 & 0.62 $\pm$ 0.26 & 0.64 $\pm$ 0.22 & 0.31 $\pm$ 0.31 & 0.55 $\pm$ 0.21 \\
\midrule
Gemini 3 Pro (preview) & -0.25 $\pm$ 0.34 & -0.85 $\pm$ 0.17 & -0.75 $\pm$ 0.18 & -1.16 $\pm$ 0.31 & -0.92 $\pm$ 0.23 & -0.57 $\pm$ 0.33 \\
\mbox{LR$\rightarrow$HR augmented NLLB} & -0.02 $\pm$ 0.24 & -0.08 $\pm$ 0.27 & -0.16 $\pm$ 0.27 & -0.01 $\pm$ 0.27 & 0.40 $\pm$ 0.26 & -0.15 $\pm$ 0.30 \\
\quad + Dictionary prompting & -0.03 $\pm$ 0.33 & 0.21 $\pm$ 0.24 & 0.36 $\pm$ 0.29 & 0.03 $\pm$ 0.26 & 0.18 $\pm$ 0.28 & 0.19 $\pm$ 0.27 \\
\bottomrule
\end{tabularx}
\normalsize

\caption{Per-variety accuracy scores for German$\rightarrow$Romansh translation. Ratings were made on the document level and then averaged across documents.}
\label{tab:detailed-human-accuracy}
\end{table}

\subsection{Pairwise Win-Rates of Systems in Human Evaluation}

\begin{table}[H]
\centering
\footnotesize
\begin{tabularx}{\columnwidth}{@{}Xrrrr@{}}
\toprule
 & \mbox{Human reference} & \mbox{Gemini 3 Pro (preview)} & \mbox{LR$\rightarrow$HR augmented NLLB} & \mbox{\quad + Dictionary prompting} \\
\midrule
\mbox{Human reference} & - & \phantom{0}72 $\pm$ \phantom{0}4 & \phantom{0}53 $\pm$ \phantom{0}4 & \phantom{0}56 $\pm$ \phantom{0}5 \\
\mbox{Gemini 3 Pro (preview)} & \phantom{00}9 $\pm$ \phantom{0}2 & - & \phantom{0}15 $\pm$ \phantom{0}3 & \phantom{0}11 $\pm$ \phantom{0}3 \\
\mbox{LR$\rightarrow$HR augmented NLLB} & \phantom{0}23 $\pm$ \phantom{0}3 & \phantom{0}63 $\pm$ \phantom{0}4 & - & \phantom{0}20 $\pm$ \phantom{0}4 \\
\mbox{\quad + Dictionary prompting} & \phantom{0}22 $\pm$ \phantom{0}4 & \phantom{0}66 $\pm$ \phantom{0}4 & \phantom{0}22 $\pm$ \phantom{0}4 & - \\
\bottomrule
\end{tabularx}
\normalsize
\caption{Win-rates of systems in segment-level fluency evaluation, on average across varieties. The values indicate the percentage of segments where the system in the row was rated higher than the system in the column.}
\end{table}

\begin{table}[H]
\centering
\footnotesize
\begin{tabularx}{\columnwidth}{@{}Xrrrr@{}}
\toprule
 & \mbox{Human reference} & \mbox{Gemini 3 Pro (preview)} & \mbox{LR$\rightarrow$HR augmented NLLB} & \mbox{\quad + Dictionary prompting} \\
\midrule
\mbox{Human reference} & - & \phantom{0}80 $\pm$ 10 & \phantom{0}66 $\pm$ 10 & \phantom{0}50 $\pm$ 14 \\
\mbox{Gemini 3 Pro (preview)} & \phantom{00}5 $\pm$ \phantom{0}4 & - & \phantom{00}4 $\pm$ \phantom{0}4 & \phantom{0}11 $\pm$ \phantom{0}5 \\
\mbox{LR$\rightarrow$HR augmented NLLB} & \phantom{0}11 $\pm$ \phantom{0}7 & \phantom{0}64 $\pm$ 10 & - & \phantom{0}26 $\pm$ \phantom{0}9 \\
\mbox{\quad + Dictionary prompting} & \phantom{0}15 $\pm$ \phantom{0}8 & \phantom{0}76 $\pm$ 12 & \phantom{0}25 $\pm$ \phantom{0}9 & - \\
\bottomrule
\end{tabularx}
\normalsize
\caption{Win-rates of systems in document-level accuracy ranking, on average across varieties.}
\end{table}

\begin{table}[H]
\centering
\footnotesize
\begin{tabularx}{\columnwidth}{@{}Xrrrr@{}}
\toprule
 & \mbox{Human reference} & \mbox{Gemini 3 Pro (preview)} & \mbox{LR$\rightarrow$HR augmented NLLB} & \mbox{\quad + Dictionary prompting} \\
\midrule
\mbox{Human reference} & - & \phantom{0}60 $\pm$ \phantom{0}5 & \phantom{0}51 $\pm$ \phantom{0}4 & \phantom{0}51 $\pm$ \phantom{0}7 \\
\mbox{Gemini 3 Pro (preview)} & \phantom{0}10 $\pm$ \phantom{0}3 & - & \phantom{0}14 $\pm$ \phantom{0}3 & \phantom{0}14 $\pm$ \phantom{0}3 \\
\mbox{LR$\rightarrow$HR augmented NLLB} & \phantom{0}20 $\pm$ \phantom{0}3 & \phantom{0}54 $\pm$ \phantom{0}4 & - & \phantom{0}19 $\pm$ \phantom{0}3 \\
\mbox{\quad + Dictionary prompting} & \phantom{0}24 $\pm$ \phantom{0}6 & \phantom{0}56 $\pm$ \phantom{0}4 & \phantom{0}22 $\pm$ \phantom{0}3 & - \\
\bottomrule
\end{tabularx}
\normalsize
\caption{Win-rates of systems in segment-level accuracy evaluation, on average across varieties.}
\end{table}

\pagebreak

\section{Human Evaluation Statistics}

\subsection{Number of Ratings}\label{app:human-evaluation-statistics}

\begin{table}[H]
\centering

\footnotesize
\begin{tabularx}{\columnwidth}{@{}Xrrrr@{}}
\toprule
Variety & Human reference & Gemini 3 Pro (preview) & LR$\rightarrow$HR augmentation & + Dictionary prompting \\
\midrule
\textbf{Rumantsch Grischun} & \textbf{266 / 38 / \phantom{0}92} & \textbf{280 / 54 / 149} & \textbf{292 / 40 / 166} & \textbf{279 / 36 / 124} \\
\hspace{1em}Rater 1 & 243 / 33 / \phantom{0}59 & 249 / 42 / \phantom{0}96 & 268 / 30 / 114 & 258 / 31 / 102 \\
\hspace{1em}Rater 2 & \phantom{0}23 / \phantom{0}5 / \phantom{0}33 & \phantom{0}31 / 12 / \phantom{0}53 & \phantom{0}24 / 10 / \phantom{0}52 & \phantom{0}21 / \phantom{0}5 / \phantom{0}22 \\
\midrule
\textbf{Sursilvan} & \textbf{250 / 46 / 188} & \textbf{260 / 49 / 176} & \textbf{245 / 54 / 179} & \textbf{260 / 55 / 140} \\
\hspace{1em}Rater 3 & 208 / \phantom{0}0 / \phantom{00}0 & 210 / \phantom{0}0 / \phantom{00}0 & 190 / \phantom{0}0 / \phantom{00}0 & 204 / \phantom{0}0 / \phantom{00}0 \\
\hspace{1em}Rater 4 & \phantom{00}0 / 37 / 112 & \phantom{00}0 / 40 / \phantom{0}96 & \phantom{00}0 / 42 / 118 & \phantom{00}0 / 43 / \phantom{0}87 \\
\hspace{1em}Rater 5 & \phantom{00}0 / \phantom{0}9 / \phantom{0}76 & \phantom{00}0 / \phantom{0}9 / \phantom{0}80 & \phantom{00}0 / 12 / \phantom{0}62 & \phantom{00}0 / 12 / \phantom{0}54 \\
\hspace{1em}Rater 6 & \phantom{0}27 / \phantom{0}0 / \phantom{00}0 & \phantom{0}31 / \phantom{0}0 / \phantom{00}0 & \phantom{0}33 / \phantom{0}0 / \phantom{00}0 & \phantom{0}36 / \phantom{0}0 / \phantom{00}0 \\
\hspace{1em}Rater 7 & \phantom{0}15 / \phantom{0}0 / \phantom{00}0 & \phantom{0}19 / \phantom{0}0 / \phantom{00}0 & \phantom{0}22 / \phantom{0}0 / \phantom{00}0 & \phantom{0}20 / \phantom{0}0 / \phantom{00}0 \\
\midrule
\textbf{Sutsilvan} & \textbf{250 / 36 / 180} & \textbf{231 / 40 / 184} & \textbf{232 / 47 / 196} & \textbf{274 / 43 / 144} \\
\hspace{1em}Rater 8 & \phantom{0}69 / 14 / \phantom{0}86 & \phantom{0}71 / 18 / \phantom{0}91 & \phantom{0}63 / 21 / \phantom{0}94 & \phantom{0}73 / 21 / \phantom{0}70 \\
\hspace{1em}Rater 9 & \phantom{0}54 / 22 / \phantom{0}90 & \phantom{0}27 / 22 / \phantom{0}93 & \phantom{0}48 / 26 / \phantom{0}96 & \phantom{0}57 / 22 / \phantom{0}70 \\
\hspace{1em}Rater 10 & 127 / \phantom{0}0 / \phantom{00}2 & 133 / \phantom{0}0 / \phantom{00}0 & 121 / \phantom{0}0 / \phantom{00}6 & 144 / \phantom{0}0 / \phantom{00}4 \\
\midrule
\textbf{Surmiran} & \textbf{245 / 51 / 246} & \textbf{263 / 29 / \phantom{0}98} & \textbf{280 / 48 / 118} & \textbf{229 / 40 / 206} \\
\hspace{1em}Rater 11 & 132 / 36 / 134 & 139 / 22 / \phantom{0}58 & 147 / 36 / \phantom{0}66 & 123 / 28 / 107 \\
\hspace{1em}Rater 12 & 113 / 15 / 112 & 124 / \phantom{0}7 / \phantom{0}41 & 133 / 12 / \phantom{0}52 & 106 / 12 / \phantom{0}99 \\
\midrule
\textbf{Puter} & \textbf{262 / 32 / \phantom{0}64} & \textbf{192 / 38 / 140} & \textbf{261 / 43 / \phantom{0}76} & \textbf{214 / 45 / 146} \\
\hspace{1em}Rater 13 & \phantom{0}83 / 32 / \phantom{0}58 & \phantom{0}71 / 38 / 132 & \phantom{0}81 / 43 / \phantom{0}64 & \phantom{0}69 / 45 / 130 \\
\hspace{1em}Rater 14 & 171 / \phantom{0}0 / \phantom{00}0 & 112 / \phantom{0}0 / \phantom{00}0 & 170 / \phantom{0}0 / \phantom{00}0 & 132 / \phantom{0}0 / \phantom{00}0 \\
\hspace{1em}Rater 15 & \phantom{00}8 / \phantom{0}0 / \phantom{00}5 & \phantom{00}9 / \phantom{0}0 / \phantom{00}8 & \phantom{0}10 / \phantom{0}0 / \phantom{0}13 & \phantom{0}13 / \phantom{0}0 / \phantom{0}16 \\
\midrule
\textbf{Vallader} & \textbf{212 / 37 / \phantom{0}75} & \textbf{200 / 37 / 110} & \textbf{220 / 48 / 180} & \textbf{195 / 44 / 118} \\
\hspace{1em}Rater 16 & 212 / \phantom{0}0 / \phantom{00}0 & 200 / \phantom{0}0 / \phantom{00}0 & 220 / \phantom{0}0 / \phantom{00}0 & 195 / \phantom{0}0 / \phantom{00}0 \\
\hspace{1em}Rater 17 & \phantom{00}0 / 37 / \phantom{0}75 & \phantom{00}0 / 37 / 110 & \phantom{00}0 / 48 / 180 & \phantom{00}0 / 44 / 118 \\
\bottomrule
\end{tabularx}
\normalsize

\caption{Statistics for human evaluation: number of fluency ratings of segments / number of accuracy ratings of documents / number of accuracy preferences of segment pairs.
Pairwise win-rates based on the latter are reported in Appendix~\ref{app:detailed-human-results}.}
\label{tab:human-evaluation-statistics}
\end{table}

\vfill

\pagebreak

\subsection{Detailed Inter-Rater Agreement Analysis}\label{app:inter-rater-agreement}

\begin{table}[H]
\centering
\footnotesize
\begin{tabularx}{0.8\columnwidth}{@{}Xrrrrrr@{}}
\toprule
\textbf{Metric} & \textbf{RG} & \textbf{Surs.} & \textbf{Suts.} & \textbf{Surm.} & \textbf{Puter} & \textbf{Vall.} \\
\midrule
Number of overlapping segments & 51 & 64 & 139 & 238 & 160 & (0) \\
\midrule
Item-level Pearson correlation & 52.5 & 73.2 & 82.4 & 65.9 & 64.0 & --- \\
Pairwise comparison agreement rate & 63.8 & 66.7 & 77.9 & 68.2 & 65.8 & --- \\
System-level Spearman correlation & 80.0 & 100.0 & 100.0 & 100.0 & 80.0 & --- \\
\bottomrule
\end{tabularx}
\normalsize
\caption{Inter-rater agreement metrics for segment-level fluency ratings.}
\label{tab:inter-rater-agreement-fluency}
\end{table}

\begin{table}[H]
\centering
\footnotesize
\begin{tabularx}{0.8\columnwidth}{@{}Xrrrrrr@{}}
\toprule
\textbf{Metric} & \textbf{RG} & \textbf{Surs.} & \textbf{Suts.} & \textbf{Surm.} & \textbf{Puter} & \textbf{Vall.} \\
\midrule
Number of overlapping documents & 16 & 21 & 37 & 23 & (0) & (0) \\
\midrule
Item-level Pearson correlation & 23.5 & 35.4 & 17.4 & 50.4 & --- & --- \\
Pairwise comparison agreement rate & 56.2 & 52.4 & 62.2 & 65.2 & --- & --- \\
System-level Spearman correlation & 80.0 & 40.0 & 100.0 & 80.0 & --- & --- \\
\bottomrule
\end{tabularx}
\normalsize
\caption{Inter-rater agreement metrics for document-level accuracy ratings.}
\label{tab:inter-rater-agreement-accuracy}
\end{table}

\begin{table}[H]
\centering
\footnotesize
\begin{tabularx}{0.8\columnwidth}{@{}Xrrrrrr@{}}
\toprule
\textbf{Metric} & \textbf{RG} & \textbf{Surs.} & \textbf{Suts.} & \textbf{Surm.} & \textbf{Puter} & \textbf{Vall.} \\
\midrule
Number of overlapping segments & 159 & 271 & 341 & 303 & (0) & (0) \\
\midrule
Item-level Pearson correlation & 43.0 & 45.3 & 36.1 & 44.6 & --- & --- \\
Pairwise comparison agreement rate & 56.0 & 63.5 & 52.2 & 59.4 & --- & --- \\
System-level Spearman correlation & 80.0 & 100.0 & 100.0 & 80.0 & --- & --- \\
\bottomrule
\end{tabularx}
\normalsize
\caption{Inter-rater agreement metrics for segment-level accuracy ratings.}
\label{tab:inter-rater-agreement-segment-accuracy}
\end{table}

\vfill

\pagebreak

\section{Human Evaluation Interface and Guidelines}
\subsection{Annotation Interface – Fluency}\label{app:annotation-interface-fluency}
\includegraphics[scale=0.5]{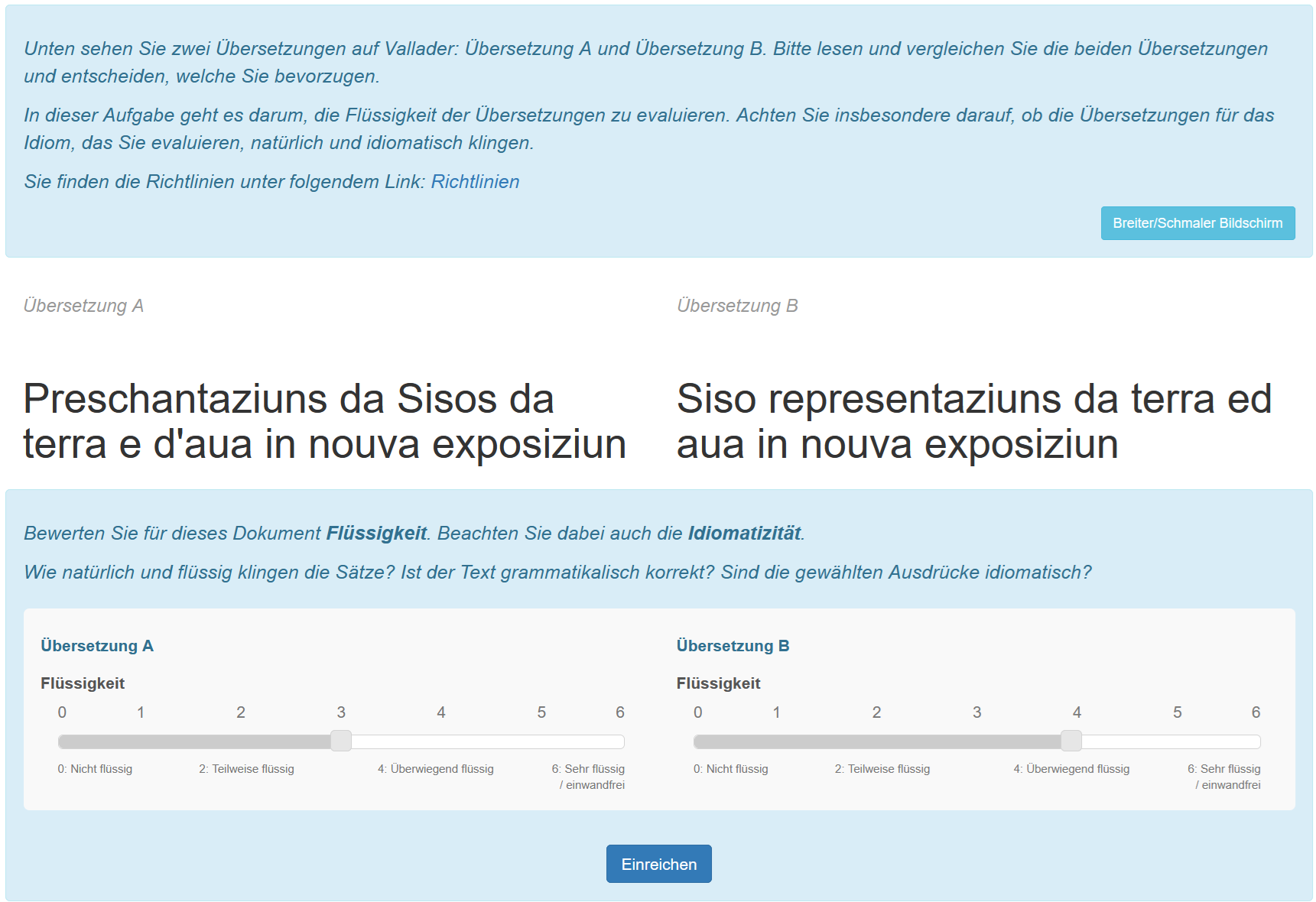}
\vfill
\clearpage

\subsection{Annotation Interface – Accuracy}\label{app:annotation-interface-accuracy}
\includegraphics[scale=0.76]{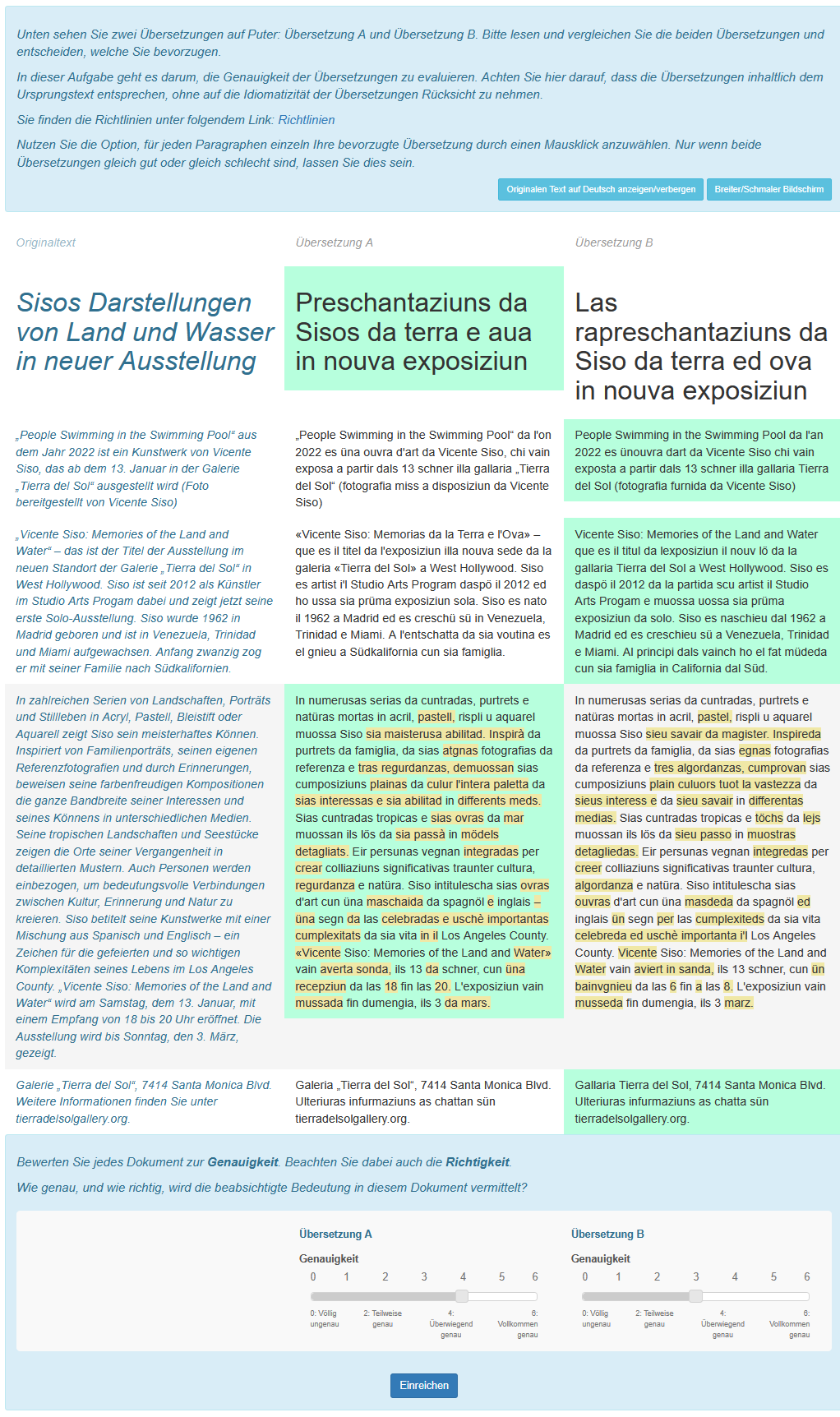}

\pagebreak

\twocolumn

\subsection{Human Evaluation Guidelines}\label{app:eval-guidelines}
\subsubsection*{General Information on the Evaluation}

\textit{For each idiom you evaluate, you will complete two passes: Fluency and Accuracy. These are explained in more detail below.
For each pass, you will receive a separate account. This means you have one account for Fluency and one account for Accuracy for each idiom you evaluate.
Click on the idiom you would like to evaluate first. Once you have started a pass, you must complete it before you can start another one. Since your two accounts exist independently of each other, it is theoretically possible not to finish the Fluency pass before starting Accuracy. However, please always finish the Fluency pass first before starting the Accuracy pass!
If you evaluate multiple idioms, we recommend evaluating both Fluency and Accuracy for the first idiom before moving on to the next idiom.}

\textit{On the evaluation page, you will now see two translations of an originally German text. Read them carefully. When evaluating Accuracy, also display the German source text. Differences between the two translations are highlighted in yellow.
More detailed guidelines for each pass can be found below.
If at any point during the evaluation process you can no longer see your task, simply copy this link: Link to platform}

\textit{Once you click ``Submit'' for a task, you can no longer access it.}

\subsubsection*{Fluency}

\textit{Here, the focus is on evaluating idiomatic usage and grammatical correctness. Pay attention to whether the translation is correct within the Romansh idiom you are evaluating—idiomatic, grammatically correct, and overall natural-sounding.
Below each segment you will find a slider from 0 to 6. Based on your own judgment, taking into account the approaches described below, estimate the score you would like to assign to each segment.
In particular, consider whether the translation corresponds to the idiom being evaluated, and deduct points accordingly for any mixing with other idioms.}

\textit{Example: You are evaluating a text in Vallader and notice that many formulations come from Puter. These are correct in Puter, but do not exist in Vallader. For the ``Fluency'' slider, choose a maximum of 2—depending on how good the remaining formulations are. If, in this example, the entire text is actually Puter, then choose 0 instead.
If idioms are mixed within a segment, evaluate it exclusively within the framework of the idiom being assessed: what is Sutsilvan in a Sursilvan text should be counted as ``incorrect.''
Invented terms or conjugations borrowed from other languages are also considered ``incorrect.'' Deduct points according to their frequency and severity.
Also evaluate grammar errors that significantly hinder comprehension more strictly.}

\subsubsection*{Accuracy}

\textit{Here you compare the two translations with the original German text to assess whether they match in terms of content.
To display the source text, click ``Show/hide original text in German'' at the top of the page.
For each segment, compare which translation is better and click on your preferred one. It will then be highlighted in green. If you find both translations equally good or equally poor, do not mark the segment.}

\textit{If a translation expresses the exact opposite of the original text, assign at least 1 point, but no more than 2.
If information is omitted or invented, deduct points accordingly.
Here as well, prefer idiomatic expressions over overly literal translations.
If the same information is present but in a different order, this does not need to be penalized as long as the meaning is not changed.
Inconsistency in the translation of a particular word—when a word is translated in several different ways—does not have to be penalized, but may be penalized with a maximum deduction of 1 point.
Untranslated proper names, provided a translation exists, likewise do not have to be penalized, but may be penalized with a maximum deduction of 1 point.}

\end{document}